\documentclass[10pt,a4paper]{article}

\usepackage{geometry}
\usepackage[english]{babel}
\usepackage{graphicx}
\usepackage{tikz}
\usepackage{framed}
\usepackage[normalem]{ulem}
\usepackage{amsmath}
\usepackage{amssymb}
\usepackage{amsfonts}

\usepackage{amsthm}
\usepackage[utf8]{inputenc}
\usepackage{caption}
\usepackage{subcaption}
\usepackage{float}
\usepackage{verbatim}
\usepackage{braket}
\usepackage{mathtools}
\usepackage{nicefrac}
\usepackage{indentfirst}
\usepackage{url}
\usepackage[colorlinks]{hyperref}
\usepackage{nameref}
\usepackage{authblk}

\hypersetup{
  allcolors = [rgb]{0.2,0.5,0.6},
  citecolor = [rgb]{0.8,0.2,0.4}
}

\newcommand{\<}{\left\langle}
\renewcommand{\>}{\right\rangle}
\newcommand{\vl}{\underline}
\newcommand{\mc}{\mathcal}
\newcommand{\mb}{\mathbb}
\newcommand{\mr}{\mathrm}
\newcommand{\tl}{\tilde}
\newcommand{\abs}[1]{\left\lvert\!\middle\lvert#1\middle\rvert\!\right\rvert}
\newcommand{\eref}[1]{Eq.~(\ref{#1})}
\newcommand{\tref}[1]{\nameref{#1}}

\newcommand{\aref}[1]{Appendix~\ref{#1}}
\newcommand{\sref}[1]{Section~\ref{#1}} 
\newcommand{\fref}[1]{Fig.~\ref{#1}}

\newcommand{\hypgeo}[2]{%
  \operatorname{%
    {\vphantom{\mathnormal{F}}}_{#1}%
    \kern-\scriptspace
    \mathnormal{F}_{#2}%
  }%
}
\newcommand{\mycite}[1]{
\cite{#1}
}

\title{How isotropic kernels perform on simple invariants}
\author[a]{Jonas Paccolat}
\author[a]{Stefano Spigler}
\author[a]{Matthieu Wyart}
\affil[a]{Institute of Physics, \'Ecole Polytechnique F\'ed\'erale de Lausanne, 1015 Lausanne, Switzerland}

\begin{document}
\maketitle

\section*{Abstract}
We investigate how the training curve of isotropic kernel methods depends on the symmetry of the task to be learned, in several settings. 
(i) We consider a regression task, where the target function is a Gaussian random field  that depends only on $d_\parallel$ variables, fewer than the input dimension $d$. We compute the expected test error $\epsilon$ that follows $\epsilon\sim p^{-\beta}$ where $p$ is the size of the training set. We find that $\beta\sim\nicefrac{1}{d}$ independently of $d_\parallel$, supporting previous findings that the presence of invariants does not resolve the curse of dimensionality for kernel regression. (ii) Next we consider support-vector binary classification and introduce the {\it stripe model} where the data label depends on a single coordinate $y(\vl x) = y(x_1)$, corresponding to parallel decision boundaries separating labels of different signs, and consider that there is no margin at these interfaces. We argue and confirm numerically that for large bandwidth, $\beta = \frac{d-1+\xi}{3d-3+\xi}$, where $\xi\in (0,2)$ is the exponent characterizing the singularity of the kernel at the origin. This estimation improves classical bounds obtainable from Rademacher complexity. In this setting there is no curse of dimensionality since $\beta\rightarrow \nicefrac13$ as $d\rightarrow\infty$.
 (iii) We confirm these findings for the {\it spherical model} for which $y(\vl x) = y(\abs{\vl x})$. (iv) In the stripe model, we show that if the data are compressed along their invariants by some factor $\lambda$ (an operation believed to take place in deep networks), the test error is reduced by a factor $\lambda^{-\frac{2(d-1)}{3d-3+\xi}}$.

\section{Introduction and related works}

Deep neural networks are successful at a variety of tasks, yet understanding why they work remains a challenge. In particular, we do not know a priori how many data are required to learn a given rule --- not even the order of magnitude. Specifically, let us denote by  $p$ the number of examples in the training set. After learning, performance is quantified by the test error $\epsilon(p)$. 
Quite remarkably, empirically one observes that $\epsilon(p)$ is often well fitted by a power-law decay $\epsilon \sim p^{-\beta}$. The exponent $\beta$ is found to depend on the task, on the dataset and on the learning algorithm (\mycite{hestness2017deep,spigler2019asymptotic}). General arguments would suggest that $\beta$ should be extremely small --- and learning thus essentially impossible --- when the dimension $D$ of the data is large, which is generally the case in practice (e.g.~in images where $D$ is the number of pixels times the number of color channels). For example in a regression task, if the only assumption on the target function is that it is Lipschitz continuous, then the test error cannot be guaranteed to decay faster than with an exponent $\beta \sim \nicefrac{1}{D}$ (\mycite{luxburg2004distance}). This \emph{curse of dimensionality} (\mycite{bach2017breaking}) stems from the geometrical fact that the distance $\delta$ among nearest-neighbor data points decays extremely slowly in large $d$ as $\delta \sim p^{\nicefrac{1}{D}}$, so that any interpolation  method is very imprecise. The mere observation that deep learning works in large dimension implies that data are very structured (\mycite{mallat2016understanding}). Yet how to describe mathematically this structure and to build a quantitative theory for $\beta$ remains a challenge.  Our present goal is to study the relationship between $\beta$ and symmetries in the data in simple models.


Recently there has been a considerable interest in studying the infinite-width limit of neural networks, 
motivated by the observation that performance generally improves with the number of parameters
(\mycite{neyshabur2017geometry,neyshabur2018towards,bansal2018minnorm,advani2017high,spigler2019jamming,geiger2020scaling}).
That limit depends on how the weights at initialization scale with the width.
For a specific choice, similar to the LeCun initialization often used in practice, deep learning becomes equivalent to a kernel method (\mycite{jacot2018neural}), which has been coined \emph{neural tangent kernel}.
In kernel methods, the learned function $Z(\vl x)$ is a linear combination of the functions $K(\vl x,\vl x_\mu)$, where $\vl x_\mu$ are the training data and $K$ is the kernel. These methods achieve performances somewhat inferior but still comparable to the more refined deep networks (\mycite{bruna2013invariant,arora2019exact}), and are often used both for regression and classification. In this work we study the learning curves of isotropic kernels for which $K(\vl x,\vl y)=K\left(\abs{\vl x-\vl y}\right)$, that include the popular Gaussian and Laplace kernels.

When these kernels are used on the  image datasets \texttt{MNIST} and \texttt{CIFAR-10}, one finds that the learning curves decay with respective exponents $\beta_\mr{MNIST}\approx0.4$ and $\beta_\mr{CIFAR-10}\approx0.1$ that are much larger than $\nicefrac{1}{D_{\mr{MNIST}}}\approx10^{-3}$ and $\nicefrac{1}{D_\mr{CIFAR10}}\approx3\cdot10^{-4}$ (\mycite{spigler2019asymptotic}). Several aspects of the data could explain together these findings that $\beta$ is much larger than $\nicefrac{1}{D}$. 
 
(i) In the kernel literature, upper bounds on the test error with $\beta$ independent of $D$ are obtained assuming that the target function lies in  the \emph{reproducing-kernel Hilbert space} of the kernel.\footnote{Such a Hilbert space is the set of all functions $f$ with finite $K$-norm: $\abs{f}_K<\infty$, see (\mycite{scholkopf2001learning}).} However for these kernels this assumption is rather extreme: it supposes that the number of derivatives  of the target function that are smooth is proportional to the dimension itself (\mycite{maiorov2006approximation,bach2017breaking}), see (\mycite{spigler2019asymptotic}) for a precise statement for Gaussian random functions.

 (ii) The data  live on a manifold $\mathfrak{M}$ of lower dimensionality $d \leq D$. This is indeed the case for \texttt{MNIST} where $d \approx 15$ (\mycite{costa2004learning,hein2005intrinsic,rozza2012novel,facco2017estimating,goldt2019modelling}) and \texttt{CIFAR-10} where $d \approx 35$ (\mycite{spigler2019asymptotic}). This effect is presumably important, yet by itself it may not be the resolution of the problem, since the exponents $\beta$ are significantly larger than $1/d$. Unless stated otherwise, in this work the data manifold extends to the whole space, namely $d=D$.
 
(iii) The function to be learned presents many invariants. It can be expressed in terms of just $d_\parallel < d$ spatial components. For example in the context of classification, some pixels at the edge of the image may be unrelated to the class label. Likewise, smooth deformations of the image may leave the class unchanged. It has been argued that the presence of these invariants is central to the success of deep learning (\mycite{mallat2016understanding}). In that view, neural networks corresponds to a succession of non-linear and linear operations where  invariant directions are compressed (\cite{paccolat2020geometric}). It is supported by the observations that kernels designed to perform such compression perform well (\mycite{mallat2016understanding})  and that compression can indeed occur at intermediate layers of deep networks (\mycite{shwartz2017opening}). Yet, relating quantitatively these views to the learning-curve exponent $\beta$ remains a challenge, even for simple isotropic kernels and simple models of data. In (\mycite{bach2017breaking}), it was shown for a specific kernel in the context of regression that the presence of invariants did not improve guaranties for  $\beta$. It is currently unclear if this results holds more generally to other kernels, beyond worst case analysis, and to classification tasks.

\subsection{Our contribution}


Our work consists of two parts that can be read independently, studying respectively regression and classification for different models.


The first part is presented in \sref{sec:regrTS} and focuses on kernel regression. We consider a target function  that varies only along a linear manifold of $d_\parallel$ directions of the input space, and is invariant along the remaining $d-d_\parallel$ directions. Without loss of generality,  we consider that this dependence is on $\vl x_\parallel \equiv (x_1,\dots,x_{d_\parallel})^t$, corresponding to the $d_\parallel$ first components of the data vectors $\vl x = (x_1,\dots,x_d)^t$. The target function is a  Gaussian random function $Z_T(\vl x) \equiv Z_T(\vl x_\parallel)$ with covariance determined by an isotropic translation-invariant Teacher kernel $K_T(\vl x)$. Kernel ridgeless regression is then performed using a distinct Student kernel $K_S(\vl x)$. Such a Teacher-Student framework (without invariants) was first introduced in (\mycite{sollich1999learning,sollich2002gaussian}) and recently generalized in (\mycite{bordelon2020spectrum}). In these references it is investigated via an approximate formula based on averaging on the randomness of the data distribution. 
Here instead we use the methods of (\mycite{spigler2019asymptotic}) inspired by earlier works on kriging (\mycite{stein1999predicting}) to compute the learning curve by calculating the expectation of the mean-squared test error, so as to extract the exponent $\beta$. We find and confirm numerically that $\beta$ is independent on $d_\parallel$ and depends only on $d$: one cannot escape the curse of dimensionality. This result supports that even in a typical, non-worst case analysis, regression using simple kernels does not benefit from invariance in the data. Beyond the dependence on $d$, the exponent $\beta$ is determined by the Teacher and Student kernels only through two exponents $\alpha_T(d),\alpha_S(d)$ related to the decay of their Fourier transforms. In \sref{sec:regrTS}, we define these exponents and we show that $\beta = \frac1d\min(\alpha_T(d)-d,2\alpha_S(d))$.


In the second part of this work, we study kernel classification with support-vector machines, for which conclusions differ. We focus on simple models of data ($d_\parallel = 1$) that are arguably necessary first steps to build quantitative predictions for $\beta$ in more realistic settings. In \sref{sec:svc}, we introduce the {\it stripe model}, in which the class label $y(\vl x)=\pm 1$ only varies in one direction, as illustrated in \fref{fig:kernelflatinterface}. Again without loss of generality, we consider $y(\vl x)=y(x_1)$. This model corresponds to parallel interfaces separating regions where the label changes sign. We further consider the case without margin, where the data distribution $\rho(\vl x)$ is non zero at interfaces.

The performance of isotropic kernel classification we focus on in this paper depends on the bandwidth $\sigma$ of the kernel, that is the scale over which it varies significantly. If $\sigma$ is much smaller than the distance $\delta$ between training points, then the support-vector machine is tantamount to a nearest-neighbor algorithm, which inevitably suffers from the curse of dimensionality with an exponent $\beta \sim \nicefrac1d$. However in the limit of large $\sigma$, we provide scaling (heuristic) arguments that we systematically confirm numerically, showing that $\beta = \frac{d-1+\xi}{3d-3+\xi}$, where $\xi$ is an exponent characterizing the singularity of the kernel at the origin (e.g. $\xi=1$ for a Laplace kernel). This exponent $\beta$ stays finite even in large dimension. 

In \sref{sec:sphere}, we show that these results are not restricted to strictly flat interfaces: the same exponent $\beta$ is found for the {\it spherical model} in which $y(\vl x) = y(\abs{\vl x})$. More generally, our analysis suggests that this result will break down  if the boundary  separating labels shows significant variation below a length scale $r_c\sim p^{-1/(d-1)}$. Avoiding the curse of dimensionality thus requires to have an increasingly regular boundary separating labels as $d$ increases.

Finally, in  \sref{sec:stretching}, we come back to the stripe model and study how compressing the input data along its invariants (namely all the directions different from $x_1$) by a factor $\lambda$ improves performance - an effect believed to play a key role in the success of deep learning (\mycite{mallat2016understanding}). We argue and confirm empirically that when mild, such a compression leaves the exponent $\beta$ unchanged but reduces the test error by a factor $\lambda^{-\frac{2(d-1)}{3d-3+\xi}}$.


\subsection{Related works}

{\it Regression:} the optimal worst-case performance of kernel regression has been investigated using a \emph{source condition} that constrains the decay of the coefficients of the true function in the eigenbasis of the covariant operator associated to the kernel (\mycite{fischer2017sobolev,caponnetto2007optimal,pillaud2018statistical}). For isotropic kernels and uniform data distribution, this condition is similar to controlling the decay of the Fourier components of the true function as we do here, and with our notation \footnote{Specifically, this literature introduces an exponent $b$ characterizing the decay of the eigenvalues $\lambda_\rho$ of the covariant operator associated to the kernel with their rank $\rho$: $\lambda_\rho\sim \rho^{-b}$. In our set-up it is straightforward to show that $b=\alpha_S/d$. Another exponent $c$ (sometimes noted $2r$ \cite{pillaud2018statistical}) characterizes the smoothness of the target function $f^\star$. It is defined as the largest exponent for which $\langle f^\star|K_S^{1-c}f^\star\rangle<\infty$. It is straightforward to show that in our case,  $c = \frac{\alpha_T - d}{\alpha_S}$. The worst case exponent  is $\beta_\mathrm{wc} = \frac{bc}{bc + 1}$ \mycite{fischer2017sobolev,caponnetto2007optimal,pillaud2018statistical} and is expressed in our notations in the main text.} the optimal worst-case generalization error is $\epsilon_\mr{wc} \lesssim p^{-\beta_\mr{wc}}$ with $\beta_\mr{wc} = \frac{\alpha_T(d)-d}{\alpha_T}$ that is independent of the Student. In addition to focusing on the worst case, these approaches generally consider noisy data - but see \cite{berthier2020tight} for recent results in the noiseless case. By contrast, in our approach the set up is noiseless. Furthermore, we average the mean square error on all Gaussian fields with a given covariance, leading to a \emph{typical} (instead of worst-case) exponent $\beta=\frac1d\min(\alpha_T(d)-d,2\alpha_S(d))$. As expected, we always have $\beta>\beta_\mr{wc}$: this follows from the fact that the exponents $\alpha_T,\alpha_S$ must be larger than $d$ for the kernels to be finite at the origin, a condition needed for our results to apply.


{\it Classification:}
There is a long history of works computing the learning curve exponent $\beta$ in regression or classification tasks
where the true function or label depends on a single direction in input space, starting from the perceptron model
(\mycite{gardner1988space}) and including support vector classification (\mycite{dietrich1999statistical}). More recently random features models have received a lot of attention, and can be analytically resolved in some cases using random matrix or replica theories (\mycite{advani2017high,montanari2019generalization,gerace2020generalisation,d2020double}).  Yet these results for classification generally consider linearly separable data\footnote{See (\mycite{dietrich1999statistical}) for an example of non-linearly separable data lying on a hypercube.}   and most importantly for both regression and classification tasks apply in the limit $d\rightarrow \infty$ and $p\rightarrow \infty$ with $\alpha=p/d$ fixed. In (\mycite{dietrich1999statistical})  for a single interface separating labels and kernels similar to ours, the learning curves of support vector classifier  was shown to decreases as $\epsilon\sim 1/\alpha$, as also found for the perceptron (\mycite{engel2001statistical}).
Here we  consider both linearly and non-linearly separable data, and take the limit of large training set size $p$ at fixed dimension $d$. It is in our view warranted considering data sets commonly used as benchmarks, such as MNIST or CIFAR for which $d_\mathfrak{M}\in [15,35]$ and $p\approx 6\cdot10^4$. In simple models for such numbers we do find that the training curves are well-described by the limit we study.  Specifically, the exponent $\beta$ we find depends  on  dimension $d$ and does not converge to $1$ as $d\rightarrow \infty$, indicating that the two limits do not commute.

Classical works on kernel classification based on Rademacher complexity lead to lower bounds on $\beta\geq 1/4$ (\mycite{bartlett2002rademacher, boucheron_bousquet_lugosi_2005}) for certain algorithms applied to the stripe and spherical model \footnote{ For example for a single interface, Theorem 21 of (\mycite{bartlett2002rademacher}) bounding the test error can be applied with a linear function $f(\vl x)=x_1$ which has a finite RKHS norm. The  bound on the test error then behaves as  $p^{-1/4}$. An algorithm minimizing the expression for the bound on all functions on the RKHS ball of identical norm must thus lead to $\beta\geq 1/4$. It is close in spirit to a SVM, supporting that the later should also satisfy $\beta\geq 1/4$. }. Our estimation thus improves on that bound, even in the limit of large dimension where we find $\beta=1/3$.

\section{Kernel regression: Teacher-Student framework}\label{sec:regrTS}

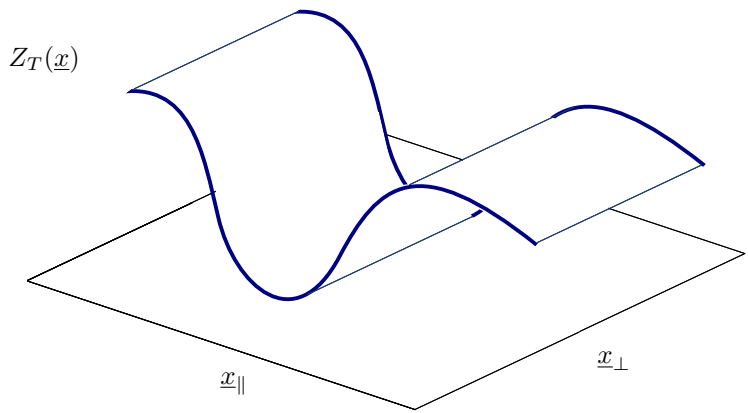
\begin{figure}[b!]
    \center
    \begin{tikzpicture}
        \node (img) {\includegraphics[scale=0.5]{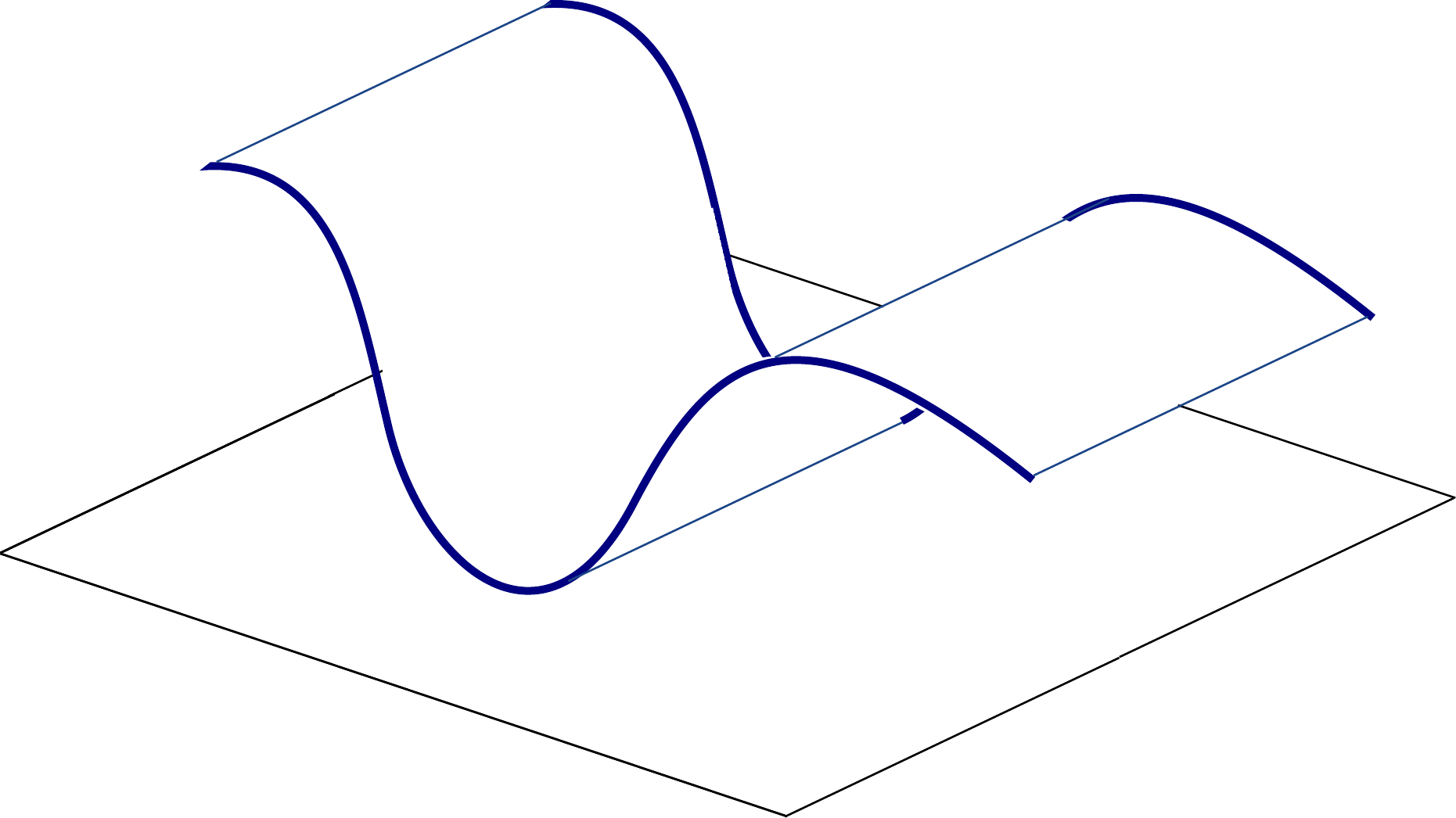}};
        \node at (3,-2) {$\vl x_\perp$};
        \node at (-2,-2.3) {$\vl x_\parallel$};
        \node at (-4.5,2) {$Z_T(\vl x)$};
    \end{tikzpicture}
    \caption{\label{fig:teacherinvariance} Sketch of a realization of the Gaussian random process $Z_T(\vl x) \sim \mc N(0,K_T)$. The kernel $K_T$, and consequently the random function $Z_T$, is constant along the direction $\vl x_\perp$ and only depends on $\vl x_\parallel$.}
\end{figure}

We consider kernel ridgeless regression on Gaussian random data that present invariants. Our framework  corresponds to a Teacher-Student setting for supervised learning (\mycite{saad1995line,monasson1995weight,opper2001advanced,engel2001statistical,
aubin2018committee,franz2018jamming}), where two variants of the same model (here kernels) are used both to generate the data and to learn them. The target function $Z_T(\vl x)$ is assumed to be a random Gaussian process $\mc N(0,K_T)$ with zero mean and covariance determined by a strictly positive-definite isotropic translation-invariant Teacher kernel $K_T(\vl x,\vl x^\prime) = K_T\left(\abs{\vl x-\vl x^\prime}\right)$, implying that $\mb E_T Z_T(\vl x) = 0$ and $\mb E_T Z_T(\vl x)Z_T(\vl x^\prime) = K_T(\vl x,\vl x^\prime)$, where we denote by $\mb E_T$ the expectation over the Teacher Gaussian random process \footnote{With respect to the kernel literature, note that in our setting $Z_T$ never belongs to the RKHS of $K_T$, see e.g. \cite{kanagawa2018gaussian}. The conditions for it to belong to $K_S$ are discussed in \cite{spigler2019asymptotic}.}.  Strictly positive-definiteness is required to generate such a random function.

We further assume that the function $Z_T(\vl x)$ does not depend on all the variables $\vl x = (x_1,\dots,x_d)^t$, but only on the first components $\vl x_\parallel \equiv (x_1,\dots,x_{d_\parallel})^t$ for some $d_\parallel \leq d$: $Z_T(\vl x) = Z_T(\vl x_\parallel)$, as sketched in \fref{fig:teacherinvariance}. The Gaussian random process $Z_T(\vl x)$ is constant along the subspace of $\vl x_\perp \equiv (x_{d_\parallel+1},\dots,x_d)^t$ when it is generated by a Teacher kernel that has the same property, namely $K_T\left(\abs{\vl x-\vl x^\prime}\right) = K_T\left(\abs{\vl x_\parallel-\vl x^\prime_\parallel}\right)$. Indeed, we have that
\begin{equation}
    \mb E_T \left[Z_T(\vl x_\parallel + \vl x_\perp) - Z_T(\vl x_\parallel)\right]^2 = 2K_T(0) - 2K_T(\vl x_\perp) = 0.
\end{equation}

The (finite) training set is made up by the values of the target function $Z_T(\vl x^\mu)$ at $p$ points $\{\vl x^\mu\}_{\mu=1}^p$.
Kernel (ridgeless) regression is performed with a Student kernel $K_S(\vl x,\vl x^\prime)$, that we also take to be isotropic and translation invariant and that can be different from the Teacher kernel $K_T(\vl x,\vl x^\prime)$. The Student has no prior knowledge of the presence of invariants: its kernel is a function of all the spatial components. 

Kernel regression consists in writing the prediction for the function $\hat Z_S(\vl x)$ at a generic point $\vl x$ as a linear combination of Student kernel overlaps on the whole training set, namely:
\begin{equation}
    \hat Z_S(\vl x) = \sum_\mu a^\mu K_S(\vl x^\mu,\vl x) \equiv \vl a \cdot \vl k_S(\vl x).
\end{equation}
The vector of coefficients $\vl a$ is determined by minimizing the mean-squared loss on the training set:
\begin{equation}
    \vl a = \mr{arg}\min_{\!\!\!\!\vl a\ \ \ \ \ } \sum_\mu \left[\hat Z_S(\vl x^\mu) - Z_T(\vl x^\mu) \right]^2.
\end{equation}
The minimization of such a quadratic loss can be carried out explicitly, and the Student prediction can be written as
\begin{equation}
    \hat Z_S(\vl x) = \vl k_S(\vl x) \cdot \mb K_S^{-1} \vl Z_T,\label{eq:regrpred}
\end{equation}
where the vector $\vl Z_T \equiv (Z_T(\vl x^\mu))_{\mu=1}^n$ contains all the samples in the training set and $\mb K_S^{\mu\nu} \equiv K_S(\vl x^\mu, \vl x^\nu)$ is the Gram matrix. By definition, the Gram matrix is always invertible for any training set if the kernel $K_S$ is strictly positive definite. The generalization error is then evaluated as the expected mean-squared error on out-of-sample data that were not used for training: numerically, it is estimated by averaging over a test set composed of $p_\mr{test}$ newly-sampled data points:
\begin{equation}
\epsilon_T = \mb E_{\vl x}\, \left[\hat Z_S(\vl x) - Z_T(\vl x)\right]^2 = \frac1{p_\mr{test}}\sum_{\mu=1}^{p_\mr{test}} \left[\hat Z_S(\vl x^\mu) - Z_T(\vl x^\mu)\right]^2.
\end{equation}
This quantity is a random variable, and we take the expectation also with respect to the Teacher process to define an average test error $\epsilon = \mb E_T \epsilon_T$ --- in the numerical simulations that we discuss later, we simply average over several runs of the Teacher Gaussian process.

We study how the expected test error $\epsilon$ decays with the size $p$ of the training set. Asymptotically for large $p$, this decay follows a power law $\epsilon \sim p^{-\beta}$. In (\mycite{spigler2019asymptotic}), $\beta$ was derived in the absence of invariants ($d_\parallel=d$), building on results from the kriging literature (\mycite{stein1999predicting}). It was found that $\beta$ depends on three quantities: the dimension $d$ and two exponents $\alpha_T(d), \alpha_S(d)$ related to the two kernels. These exponents describe how the Fourier transform of the kernels decay at large frequencies: $\tl K_T(\vl w) \sim \abs{w}^{-\alpha_T(d)}$, and similarly for the Student $K_S$. Notice that since the kernels are translation invariant, their Fourier transform is a function of only one frequency vector $\vl w$. Moreover, the exponents $\alpha_T(d), \alpha_S(d)$ depend on the dimension of the space where the Fourier transform is computed. 



Our main theorem, formally presented with its proof  in \aref{app:formalregrTSthm}, is as follows:\vspace{0.5em}

\paragraph{Theorem 1}\label{th:infintrinsicdimkerregr} $\!\!\!\!$ (Informal) \emph{Let $\epsilon$ be the average mean-squared error of the regression made with a Student kernel $K_S$ on the data generated by a Teacher kernel $K_T$, sampled at points taken on a regular $d$-dimensional square lattice in $\mb R^d$ with fixed spacing $\delta$. Assume that the Teacher kernel only varies in a lower dimensional space: $K_T(\vl x) = K_T(\vl x_\parallel)$, with $\vl x_\parallel = (x_1,\dots,x_{d_\parallel})^t$ a vector in $d_\parallel \leq d$ dimensions. The Student kernel on the contrary varies along all $d$-dimensional directions in input space. Let the Fourier transforms of the two kernels decay at high frequency with dimension-dependent exponents $\alpha_T(d)$ and $\alpha_S(d)$. Then as $\delta\to0$, $\epsilon \sim \delta^{\beta d}$ with
\begin{equation}
\beta=\frac1d\min(\alpha_T(d_\parallel) - d_\parallel, 2\alpha_S(d)).
\end{equation}}\vspace{0.5em}

\underline{Note 1}: We expect that under broad conditions the quantity $\alpha_T(d_\parallel)-d_\parallel \equiv \theta_T$ (as well as $\theta_S$ obviously) does not depend on $d_\parallel$, and that $\theta_T$ corresponds to the exponent characterizing the singular behavior of $K_T(\vl x)$ at the origin:
\begin{equation}
K_T(\vl x)= C_0 |\vl x|^{\theta_T} + \mathrm{regular \ terms}
\end{equation}
as discussed in \aref{app:formalregrTSthm}. This fact can be shown (see below) for Laplace (where $\theta_T=1$) and Mat\'ern kernels whose Fourier transform can be computed exactly. Thus we recover the curse of dimensionality since $\beta = \frac1d\min(\theta_T, 2d+2\theta_S) \leq\theta_T/d$, which is independent of $d_\parallel$ and thus of the presence of invariants.


\underline{Note 2}: A remark is in order for the case of a Gaussian kernel $K(z) = \exp\left(-z^2\right)$, since it is a smooth function and its Fourier transform (being a Gaussian function too) decays faster than any power law at high frequencies. As discussed and verified in the aforementioned paper, this Theorem applies also to Gaussian kernels, provided that the corresponding exponent is taken to be $\theta=\infty$. In particular, if the Teacher is Gaussian and the Student is not, $\beta = 2 + \frac{2\theta_S}d$; in the opposite scenario, where the Teacher is not Gaussian but the Student is, $\beta = \frac{\theta_T}d$; if both kernels are Gaussian, $\beta=\infty$ and the test error decays with respect to the training set size faster than a power law. \\

{\bf Interpretation:} The following interpretation can be given for \tref{th:infintrinsicdimkerregr} when  $\alpha_S$ is large, leading to $\beta = \frac{\theta_T}d$. An isotropic kernel corresponds to a Gaussian prior on the Fourier coefficients  of the true function being learned, a prior whose magnitude decreases with wave vectors as characterized by the exponent $\alpha_S$. Clearly, the number of coefficients that can be correctly reconstructed cannot be larger than the number of observations $p$.   For large $\alpha_S$, we find that kernel regression indeed reconstructs well a number of the order of $p$ first Fourier coefficients, which corresponds to wave vectors $\vl w$ of  norm  $\abs{\vl w} \leq \nicefrac1\delta\sim p^{1/d}$. Fourier coefficients of  larger wave vectors cannot be reconstructed however, and the mean square error is then simply of order of the sum of the squares of these coefficients: 
\begin{equation}
\label{int}
    \epsilon \sim \int_{\abs{\vl w} \geq  p^{1/d}} d^{d_\parallel}\vl w \, \abs{\vl w}^{-\alpha(d_\parallel)}\sim p^{[d_\parallel -\alpha(d_\parallel)]/d}\sim p^{-\theta_T/d}.
\end{equation}

\begin{figure}[t!]
    \center
    \includegraphics[scale=0.65]{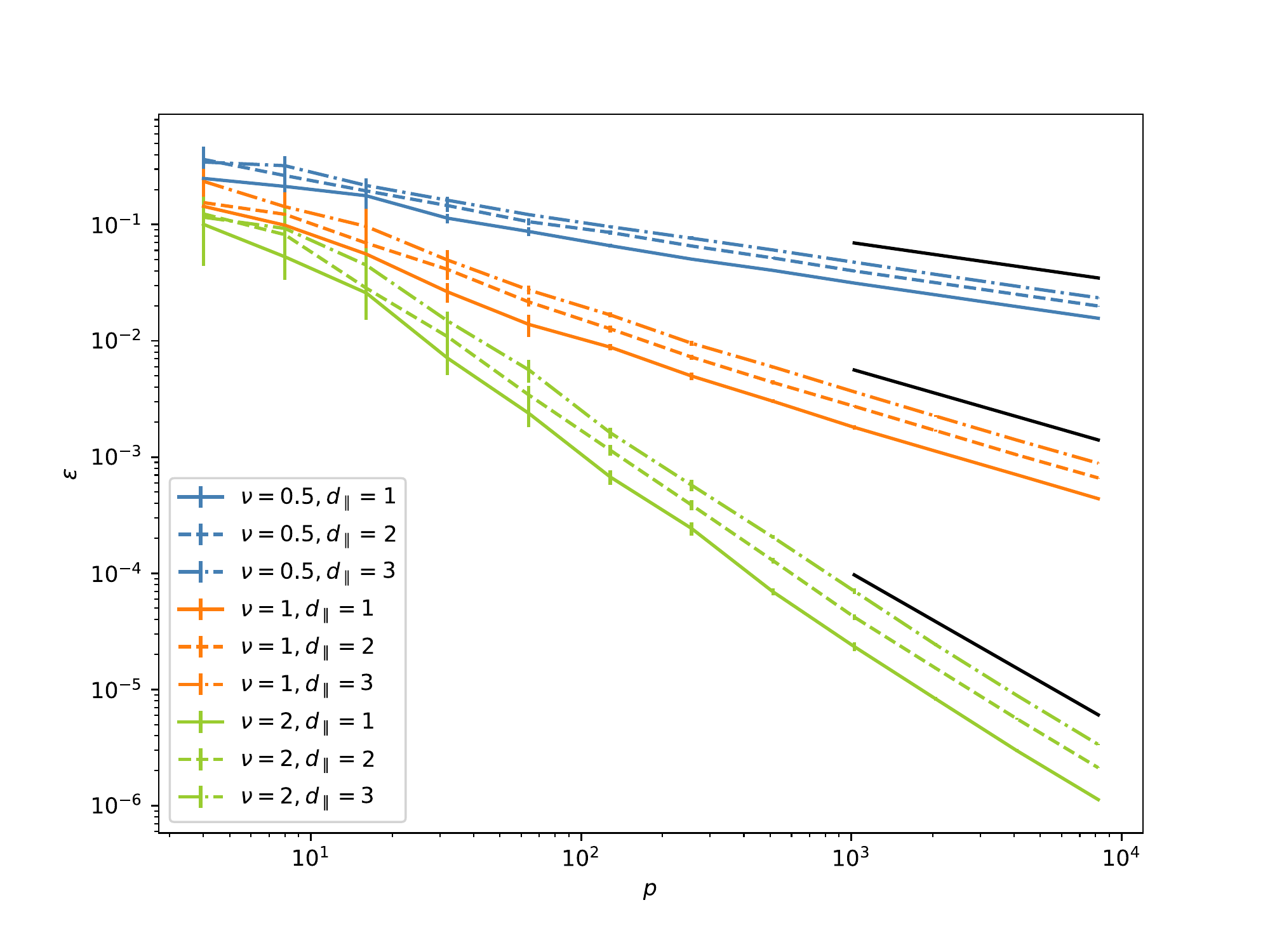}
    \caption{\label{fig:intrinsicdimkerregr} Test error $\epsilon$ {\it vs} the size $p$ of training set for Gaussian data with Matérn covariance regressed using a Laplace kernel. Identical  colors correspond to the same parameter $\nu$ of the Teacher Matérn kernel but varying dimension $d_\parallel$ as indicated in the legend.  $d_\parallel$ has no effect on the exponent $\beta$. The solid black lines represent the predicted power law with exponent $\beta = \frac23\min(\nu,4)$.}
\end{figure}

{\bf Numerical Test:}

We now test  numerically that kernel regression is blind to the lower-dimensional nature of the task. We consider a $d = D-1$-dimensional sphere of unit radius $\mb S^d$ embedded in $\mb R^D$. To test robustness with respect to our technical assumption of data points lying on an infinite lattice, we consider instead $p$ i.i.d.~points sampled uniformly at random.
The component $x^\mu_i$ of each point is generated as a standard Gaussian $\mc N(0,1)$ and then the vector $\vl x^\mu$ is normalized by dividing it by its norm. Points belonging to such a training set have a typical nearest-neighbor distance $\delta \sim p^{-1/d}$, and we will show that the test error decays with the predicted scaling $\epsilon \sim \delta^{\beta d} = p^{-\beta}$. For the numerical verification we take the Student to be a Laplace kernel:
\begin{equation}
    K_S(z) = \exp\left(-\frac{z}{\sigma}\right),
\end{equation}
that is characterized by $\alpha(d) = d + \theta_S$ with $\theta_S=1$. As Teacher we use Mat\'ern kernels, which are a family of kernels parametrized by one parameter $\nu$:
\begin{equation}\label{eq:matern_kernel}
    K_{T,\nu}(z) = \frac{2^{1-\nu}}{\Gamma(\nu)} \left(\sqrt{2\nu} \frac{z}{\sigma}\right)^\nu \mc K_\nu\left(\sqrt{2\nu} \frac{z}{\sigma}\right),
\end{equation}
where $\mc K_\nu(z)$ is the modified Bessel function of the second kind with parameter $\nu$, and $\Gamma$ is the Gamma function. Varying $\nu$ one can change the smoothness of the instances of the Gaussian random process, and in particular $\alpha_T(d) = d + \theta_T$ with $\theta_T=2\nu$. 

In the simulations, we set the spatial dimension to $D=4$ and we vary the amount of invariants in the task by taking $d_\parallel=1,2,3$. In order to fix $d_\parallel$ we simply use $z = \abs{\vl x_\parallel-\vl x^\prime_\parallel}$ instead of $z=\abs{\vl x-\vl x^\prime}$ when computing the Teacher kernel. The scale of the kernel is fixed by the constant $\sigma$, that we have taken equal to $4$ for both the Teacher and the Student. Notice that in \tref{th:infintrinsicdimkerregr} the value of $\sigma$ does not play any role since it does not enter in the asymptotic behavior of the test error (at leading order). In \fref{fig:intrinsicdimkerregr} we show that the numerical simulations match our predictions. Indeed, in this specific case the predicted exponent is
\begin{equation}
    \beta = \frac23\min(\nu,4),
\end{equation}
Notice that the exponent that characterizes the learning curves is indeed independent of $d_\parallel$. Its prefactor may however depend on  $d_\parallel$ in general.

\section{Support Vector Classification and stripe model}\label{sec:svc}

\subsection{The stripe model}
\begin{figure}[t]
    \centering
    \begin{tikzpicture}
        \node (img) {\includegraphics[scale=0.58]{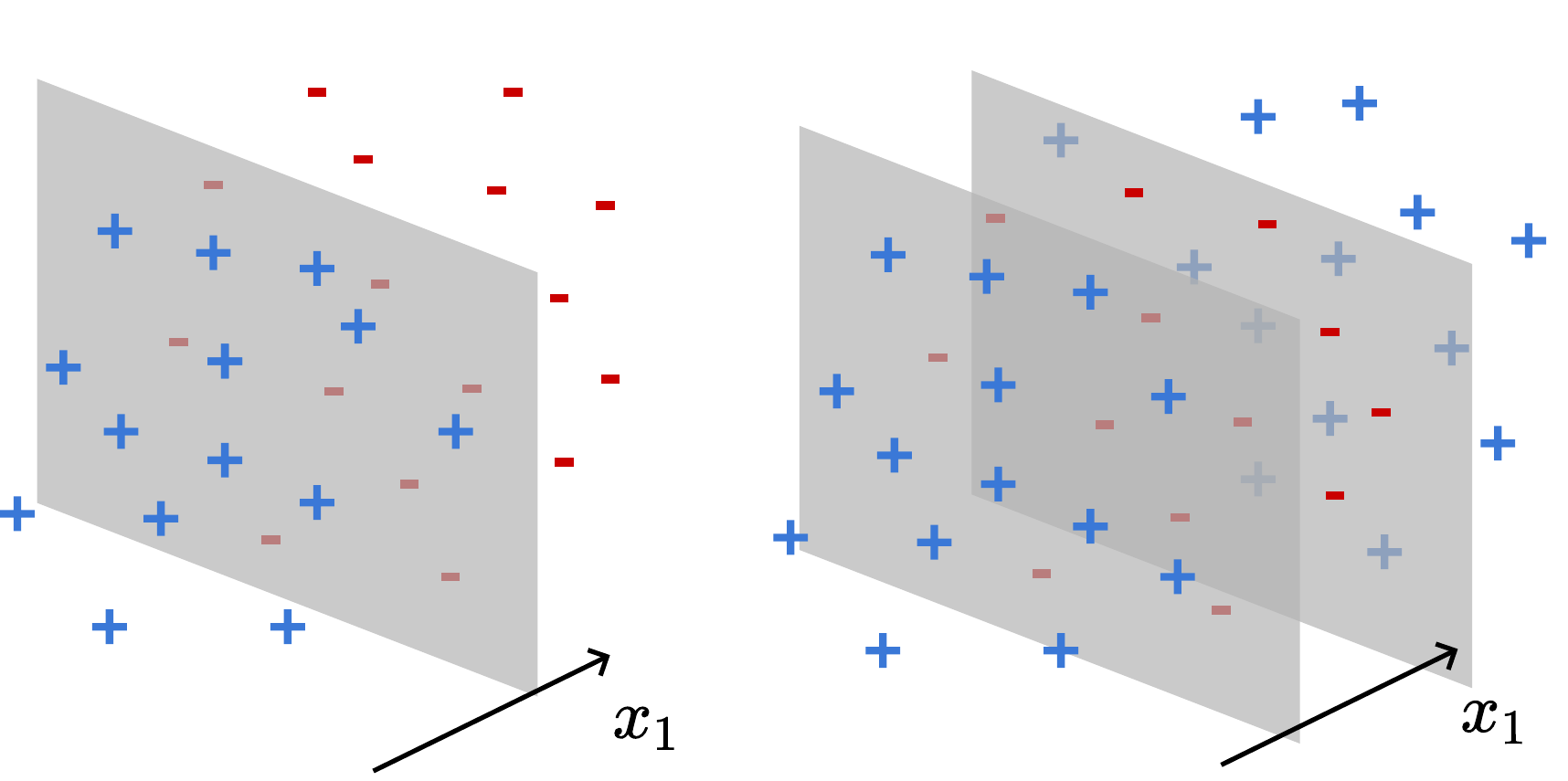}};
        \node at (-6,2) {$y(\vl x)$};
    \end{tikzpicture}
    \caption{\label{fig:kernelflatinterface} Example of decision boundaries considered in the stripe model, where the label $y(\vl x)$ of a point $\vl x$ depends only on its first component $x_1$. On the left is the \textbf{single-interface setup} where the label function is $y=+1$ on one side of the interface and $y=-1$ on the other. Points labeled in such a way compose a linearly separable dataset. On the right is the \textbf{double-interface setup}, where points are labeled $y=-1$ in between the two parallel hyperplanes and $y=+1$ on the outside.}
\end{figure}

We consider a binary classification task where the labels depend only on one direction in the data space, namely with $y(\vl x)=y(x_1)$. Layers of $y=+1$ and $y=-1$ regions alternate along the direction $x_1$, separated by parallel planes. Two examples of this setting are sketched in \fref{fig:kernelflatinterface}, corresponding to a single and double interface.  The points $\vl x$ that constitute the training and test set are iid of distribution $\rho(\vl x)$. To lighten the notation, we assume that $\rho(\vl x)$ is uniform on a square box $\Omega$ of  linear extension $\gamma$. Yet we expect our arguments to apply more generally if $\rho(\vl x)$ is continuous and does not vanish at the location of the interfaces (no margin). To confirm this view we will test and confirm below our predictions when $\rho(\vl x)$ is Gaussian distributed, with each component $x_i \sim \mc N(0,\gamma^2)$ with some variance $\gamma^2$.

\subsection{Definition of margin SVC}\label{sec:margin_SVC}

In this section we consider margin support-vector classification (margin SVC). This algorithm maximizes the margin between a decision boundary and the points in the training set that are closest to it. The prediction of the label $\hat y(\vl x)$ of a new point $\vl x$ is then made according to the sign of the estimated decision function (\mycite{scholkopf2001learning}):
\begin{equation}
f(\vl x) = \sum_{\mu=1}^p \alpha^{\mu} y^{\mu} K\left(\frac{\abs{\vl x^\mu-\vl x}}{\sigma}\right) + b \:\:\: \longrightarrow \:\:\: \hat y(\vl x) = \mr{sign}\,f(\vl x),\label{eq:svdecbound}
\end{equation}
where the kernel $K$ is conditionally strictly positive definite (\mycite{smola1998connection}) --- a condition defined in \aref{app:proof_cspd}, less stringent than strictly positive definite. In \eref{eq:svdecbound} we write explicitly the kernel bandwidth $\sigma$ since it will soon play an important role. The formulation of the margin-SVC algorithm presented below is what is referred to as the \emph{dual formulation}, but it can be equivalently recast as an attempt to maximize a (signed) distance between training points and the decision boundary (\mycite{scholkopf2001learning}).
In this dual formulation, the variables $\alpha^\mu$ are fixed by maximizing
\begin{equation}\label{eq:SVC_max_equation}
    \max_{\vl \alpha} \mc L(\vl \alpha), \ \text{with} \ \mc L(\vl \alpha) = \sum_{\mu=1}^p \alpha^\mu - \frac12 \sum_{\mu,\nu=1}^p \alpha^\mu\alpha^\nu y^\mu y^\nu K\left(\frac{\abs{\vl x^\mu-\vl x^\nu}}{\sigma}\right),
\end{equation}
subject to the constraints
\begin{align}
    \alpha^\mu \geq 0,\\
    \alpha^\mu > 0 \ \text{if and only if} \ y^\mu f(\vl x^\mu) = 1,\label{eq:svdef}\\
    Q \equiv \sum_{\mu=1}^p \alpha^\mu y^\mu = 0 \quad \text{(charge conservation)}\label{eq:chcons}.
\end{align}
The bias $b$ is set to satisfy
\begin{equation}
    \mr{min}_{1\leq\mu\leq p} |f(\vl x^{\mu})| = 1 \quad \text{(canonical condition)}\label{eq:cancond}.
\end{equation}
\eref{eq:svdef} states that a dual variable $\alpha^\mu$ is strictly positive if and only if its associated vector $\vl x^\mu$ lies on the margin, that is $y^\mu f(\vl x^{\mu}) = 1$, otherwise it is zero. Vectors with $\alpha^\mu>0$ are called \emph{support vectors} (SVs) and are the only ones that enter in the expansion of the decision function \eref{eq:svdecbound}.

\subsection{Some limiting cases of SVC}\label{sec:limiting_SVC_cases} 

{\bf Vanishing bandwidth:}
If the kernel function $K(z)$ decreases exponentially fast with some power of $z$, then in the limit $\sigma \ll \delta$, where $\delta$ is the average nearest-neighbor distance in the training set, the support-vector machine becomes akin to a nearest-neighbor algorithm. A detailed analysis of this regime for the stripe model is presented in \aref{app:smallsigma}, here we provide a qualitative argument assuming that the bias $b$ is negligible. If so, as $\sigma\rightarrow0$ one has for any training point that  $f(\vl x^\mu)\approx \alpha^\mu y^\mu K(0)$, implying that $\alpha^\mu\neq 0$ to satisfy $|f(\vl x^\mu)|\geq1$: every point is a support vector with identical $\alpha^\mu$. $f(\vl x)$ at a random test point $\vl x$ is dominated by the closest support vector. The classification error is susceptible to the curse of dimensionality for such an algorithm, and one expects generically $\epsilon \sim \delta\sim p^{-\nicefrac{1}{d}}$, as tested numerically in \fref{fig:small_sigma} for the stripe model.

{\bf Diverging bandwidth:}
In this work we focus on the other extreme case where the bandwidth is larger than the system size, namely $\sigma \gg \gamma$. In this regime the kernel is always evaluated close to the origin. Assuming that the kernel has a finite derivative in the neighborhood of the origin, we approximate it by its truncated Taylor expansion:
\begin{equation}
    K\left(\frac{\abs{\vl x-\vl x^\prime}}\sigma\right) \approx K(0) - \mr{const}\cdot \left(\frac{\abs{\vl x-\vl x^\prime}}\sigma\right)^{\xi} + o\left((\gamma/\sigma)^\xi\right).\label{eq:kertaylor}
\end{equation}
The exponent $\xi$ is related to the exponent $\theta$ introduced in \sref{sec:regrTS} by $\xi=\min(\theta,2)$, and varies from kernel to kernel. For instance, we have $\xi=1$ for Laplace kernels, $\xi=2$ for Gaussian kernels, $\xi=\tilde\gamma$ for $\tilde\gamma$-exponential kernels\footnote{We use $\tilde\gamma$ to distinguish it from the variance of the data points.} and $\xi=\min(2\nu,2)$ for Matérn kernels. In \aref{app:proof_cspd} we show that for $0<\xi<2$ the right-hand side is conditionally strictly positive definite (CSPD), which is the necessary condition for the SVC algorithm to converge. In what follows, we consider $0<\xi<2$, which excludes the Gaussian case. A proof that in that case the margin-SVC algorithm with the truncated kernel in \eref{eq:kertaylor} leads to the same solution as with the full kernel in the limit $\sigma \gg \gamma$ is presented in \aref{app:infsigmalimit}. Also, due to the charge conservation in \eref{eq:chcons}, the constant term $K(0)$ in \eref{eq:kertaylor} may safely be ignored. 

The decision function \eref{eq:svdecbound} associated to the considered radial power kernel hence becomes
\begin{equation}
    f(\vl x) =  b - \sum_{\mu=1}^p \alpha^\mu y^\mu \left(\frac{\abs{\vl x - \vl x^\mu}}{\sigma}\right)^\xi.\label{eq:longdistdecfunc}
\end{equation}
where the positive constant in \eref{eq:kertaylor} has been removed by rescaling the bias and the $\alpha^\mu$.

\subsection{ Single interface}\label{sec:large_sigma}
We consider a single interface at location $x_1=0$, with negative labels for $x_1<0$ and positive ones for $x_1>0$. Already in that case, computing analytically the test error remains a challenge, and we resort to a scaling (asymptotic) analysis to compute $\beta$. As $p$ increases, support vectors will be present on a narrower and narrower  band around the interface. We denote by $\Delta$ the characteristic extension  of that band.  $\Delta$ will depend in general on the position $\vl x_\perp$ along the interface. Here we will not study this dependence, as we are  interested on its asymptotic behavior with $p$, $\gamma$ and $\sigma$ and only track how quantities depend on these variables. From the canonical condition \eref{eq:cancond} of support vectors we have that the function $f$ varies of order one from one side of the band to the other:
\begin{equation}
\label{01}
f(\vl x_\perp +\Delta \vl e_1)-f(\vl x_\perp -\Delta \vl e_1)\sim 1, 
\end{equation}
where $\vl e_1$ is the unit vector orthogonal to the interface and $\vl x_\perp$ is any vector parallel to the plane. 

\begin{figure}[ht]
    \centerline{\includegraphics[scale=0.46]{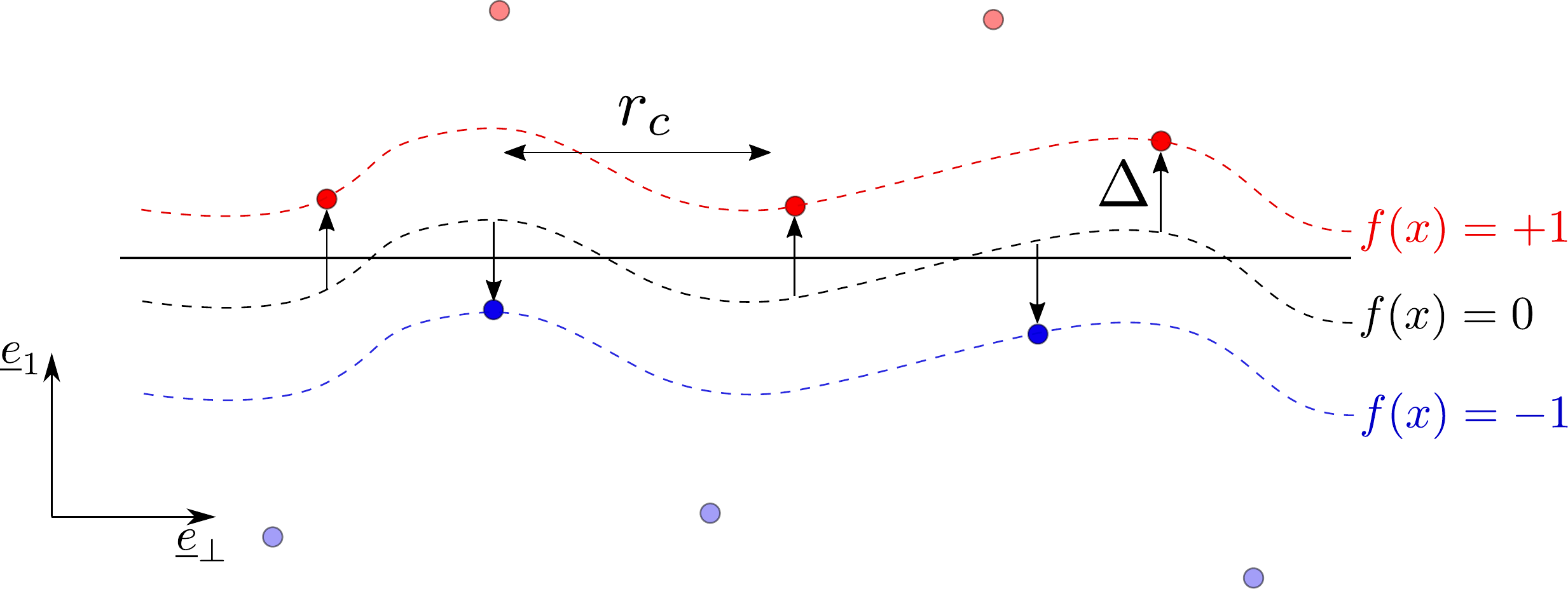}}
    \caption{\label{fig:single_interface} Sketch of the decision boundary along the interface of the stripe model. Positive and negative points are respectively represented in red and blue ; the dark points correspond to the support vectors. The predicted decision boundary (dashed black line) oscillates around the true decision boundary (solid black line) with a wave length of the order of $r_c$, the distance between nearest support vectors. The characteristic distance between support vectors and the decision line $\Delta$ is much smaller than $r_c$.}
\end{figure}

Another useful quantity is the distance $r_c$ between nearest support vectors. It can be estimated by counting the number of points lying within a cylinder of height $\Delta$ (along $x_1$) and radius $r_c$ centered on a SV, whose volume follows $\sim\Delta r_c^{d-1} $. Using that the density of data points is $\sim p/\gamma^d$, and imposing that the cylinder contains only one additional SV yields our first scaling relation:
\begin{equation}
\label{eq:scaling1}
    \frac {p}{\gamma^d} \cdot \Delta r_c^{d-1} \sim 1 \quad\Longrightarrow\quad \boxed{p\Delta r_c^{d-1}\sim \gamma^d.}
\end{equation}

Finally, the last scaling relation results from the function fluctuations being of order one within the band of support vectors when moving parallel to the true boundary decision. Indeed, we shall show below that  the function gradient along $\vl e_1$ is constant at leading order in $\Delta$, and of order $1/\Delta$ following Eq.\ref{01}.
Then the facts that (i)  on each SV the function is fixed by $f(\vl x^\mu)=y^\mu$ and (ii) the distance of the SV with respect to the true boundary fluctuates by a characteristic distance $\Delta$ jointly imply that the fluctuations  of $f(\vl x^\mu)$ as $\vl x^\mu$ evolves along the true boundary decision must be of order one. This effect is illustrated in Fig.\ref{fig:single_interface}. The characteristic transverse displacement along which these fluctuations decorrelate is simply the distance among support vector $r_c$, thus:

\begin{equation}
\label{03}
f(\vl x_\perp +r_c \vl e_\perp)-f(\vl x_\perp)\sim 1,
\end{equation}
where $\vl e_\perp$ is any unit vector parallel to the plane. Due to these fluctuations, test points inside the band have a finite probability to be incorrectly classified, and at fixed $d$ \footnote{ The value of $f(\vl x)$ in the band is governed by the neighboring support vectors, whose characteristic number is independent of $p$ but should grow with $d$. We believe  this effect to be responsible for the non-commutativity of the limits $\lim_{p/d\rightarrow\infty}\lim_{d\rightarrow\infty}$ and $\lim_{d\rightarrow\infty}\lim_{p\rightarrow\infty}$. Indeed in the former case, it is found (\mycite{dietrich1999statistical}) that $\epsilon$ and the fraction of support vectors $\Delta/\gamma$ scale differently with $\alpha$, unlike what we argue and confirm numerically in the second limit. We have checked numerically (not shown) that the ratio of these two quantities is indeed decaying with $d$ at fixed $p$.}
the test error must be proportional to the fraction $ \Delta/\gamma $ of points falling in that band:
\begin{equation*}
\label{04}
\epsilon\sim \Delta/\gamma.
\end{equation*}
We now show that from these considerations alone $\beta$ can be computed. 
Starting from \eref{eq:longdistdecfunc} we estimate the gradient of $f$ along the normal direction $\vl e_1$ at any point on the interface:
\begin{equation}
    \partial_{x_1} f(\vl x_\perp) = \frac{\xi}{\sigma} \sum_{\mu\in\Omega_{\Delta}} \alpha^\mu y^\mu \left(\frac{\abs{\vl x_\perp-\vl x^\mu}}\sigma\right)^{\xi-1} \frac{ x_1^\mu}{\abs{\vl x_\perp-\vl x^\mu}}
    \approx \xi \sigma^{-\xi} p \frac{\Delta}{\gamma}\<\alpha^\mu y^\mu x_1^\mu \abs{\vl x_\perp-\vl x^\mu}^{\xi-2}\>_{\mu\in\Omega_\Delta}\!\!\!\!\!\!\!\!\!\!\!,\label{eq:svcgradient}
\end{equation}
where the sum is over all SVs $\vl x_\mu$ indicated by the set $\Omega_\Delta$. The sum is replaced by its central-limit theorem value valid for large $p$, and we use that the number of terms in that sum goes as $p\Delta/\gamma$. The average in \eref{eq:svcgradient} scales as $\bar \alpha \Delta \gamma^{\xi-2}$ where $\bar \alpha$ is the mean value of the dual variables $\alpha^\mu$. Imposing that $\Delta \partial_{x_1} f(\vl x_\perp) \sim 1$ as follows from \eref{01} then leads to our second scaling relation:
\begin{equation}
\label{eq:scaling2}
    \boxed{p \, \bar\alpha \, \left(\frac{\Delta}{\gamma}\right)^3 \sim \left(\frac{\sigma}{\gamma}\right)^\xi.}
\end{equation}

Next we compute the consequences of \eref{03}, by recasting it in a more suitable format. We define a smoothed function $\bar f(\vl x_\perp)$ of $f(\vl x_\perp)$ on a scale $r_c$:
\begin{equation}
    \bar f(\vl x_\perp) = \int\mr d^{d-1}\vl x_\perp^\prime\, f(
    \vl x_\perp^\prime)\,G(\vl x_\perp - \vl x_\perp^\prime),\label{eq:smoothened}
\end{equation}
where the function $G$ is the Fourier transform of $\theta(\nicefrac{1}{r_c} - \abs{\vl k_\perp})$ (which is thus small when $\abs{\vl x_\perp - \vl x_\perp^\prime}\gg r_c$): 
\begin{equation}
    G(\vl x_\perp) = \int_{\abs{k_\perp}<\nicefrac{1}{r_c}}\mr d^{d-1}\vl k_\perp e^{-i\vl k_\perp\cdot x_\perp}.
\end{equation}
Thus $\bar f(\vl x_\perp)$ is obtained by removing from $f(\vl x_\perp)$ the Fourier components $\abs{k_\perp}>\nicefrac{1}{r_c}$.
The constraint of \eref{03} is equivalent to imposing that the fluctuations between $f$ and $\bar f$ are of order one.
Integrated on space it means that:
\begin{equation}\label{eq:fluctuation_over_space}
    \gamma^{-d+1}\int\mr d^{d-1}\vl x_\perp \left[ f(\vl x_\perp) - \bar f(\vl x_\perp)\right]^2 \sim 1,
\end{equation}
that can be Fourier-transformed as:
\begin{equation}
    \int\mr d^{d-1}\vl k_\perp \left[ \tl f(\vl k_\perp) - \tl f(\vl k_\perp)\,\tl G(\vl k_\perp)\right]^2 = \int_{\abs{\vl k_\perp}>\nicefrac{1}{r_c}}\mr d^{d-1}\vl k_\perp \tl f^2(\vl k_\perp)  \sim \gamma^{d-1}.\label{eq:boundedfluctuations}
\end{equation}
The Fourier transform of the decision function along the transverse components can be computed as
\begin{equation}\label{eq:decision_function_fourier_transform}
    \tl f\left(\vl k_\perp\right) = \int \mr{d}^{d-1}\vl x_\perp\, e^{-i \vl k_\perp \cdot \vl x_\perp} f(\vl x_\perp)   = \sum_{\mu\in\Omega_\Delta} \alpha^\mu y^\mu \int \mr{d}^{d-1}\vl x_\perp\, e^{-i \vl k_\perp \cdot \vl x_\perp} K\left(\frac{\abs{\vl x^\mu-\vl x_\perp}}{\sigma}\right).
\end{equation}
Using that $\abs{\vl x^\mu-\vl x_\perp}\approx \abs{\vl x_\perp^\mu-\vl x_\perp}$ and changing variables one obtains
\begin{equation}\label{eq:f_fourier_transform}
  \tl f\left(\vl k_\perp\right)   \approx \sum_{\mu\in\Omega_\Delta} \alpha^\mu y^\mu e^{-i \vl k_\perp \cdot \vl x_\perp^\mu} \cdot \int \mr{d}^{d-1}\vl x_\perp\, e^{-i \vl k_\perp \cdot \vl x_\perp} K\left(\frac{\abs{\vl x_\perp}}{\sigma}\right) \equiv \tl Q\left(\vl k_\perp\right) \cdot \tl K_\perp\left(\vl k_\perp\right),
\end{equation}
where we have defined the kernel (transverse) Fourier transform $\tl K_\perp\left(\vl k_\perp\right)$ and the "charge" structure factor $\tl Q\left(\vl k_\perp\right)$. The former can be readily computed for Laplace and Mat\'ern kernels, and at large frequencies it behaves as $\tl K_\perp\left(\vl k_\perp\right)\sim \sigma^{-\xi} \, \abs{\vl k_\perp}^{-(d-1+\xi)}$. Concerning the charge structure factor, for $\abs{\vl k_\perp}\gg\nicefrac{1}{r_c}$, the phases associated to each term in the sum defining it vary significantly even between neighboring SVs. From a central-limit argument the factor $\tl Q$ then tends to a random variable with 0 mean and variance $\bar\alpha^2 p \Delta/\gamma$. It is verified in \aref{app:charge_structure}.

We can now estimate the integral in \eref{eq:boundedfluctuations}:
\begin{equation}
    \int_{\abs{\vl k_\perp} > \nicefrac{1}{r_c}} \mr{d}^{d-1}\vl k_\perp\, \tl f^2\left(\vl k_\perp\right) \sim
    \bar\alpha^2 p \frac{\Delta}{\gamma} \sigma^{-2\xi} \int_{\abs{\vl k_\perp} > k_c} \mr{d}^{d-1}\vl k_\perp\, \abs{\vl k_\perp}^{-2(d-1+\xi)}
    \sim \bar\alpha^2 p \frac{\Delta}{\gamma} \sigma^{-2\xi} r_c^{d-1+2\xi}.
\end{equation}
The condition \eref{eq:boundedfluctuations} leads to the last scaling relation:
\begin{equation}\label{eq:scaling3}
    \boxed{\bar\alpha^2 p \frac{\Delta}{\gamma} \left(\frac{r_c}{\gamma}\right)^{d-1+2\xi} \sim \left(\frac{\sigma}{\gamma}\right)^{2\xi}.}
\end{equation}

Putting all the scaling relations together we find:
\begin{equation}\label{eq:power_laws}
    \boxed{\Delta \sim \gamma \, p^{-\frac{d-1+\xi}{3d-3+\xi}}, \quad\quad \bar\alpha \sim \left(\frac{\sigma}{\gamma}\right)^\xi \, p^{\frac{2\xi}{3d-3+\xi}}, \quad\quad r_c \sim \gamma \, p^{-\frac{2}{3d-3+\xi}}.}
\end{equation}
And consequently the asymptotic behavior of the test error is given by
\begin{equation}
  \boxed{  \epsilon \sim \frac{\Delta}{\gamma} \sim p^{-\beta}, \quad\quad \text{with } \beta = \frac{d-1+\xi}{3d-3+\xi}.\label{eq:scalelargesigma}}
\end{equation}

\underline{Note 1}: The second scaling argument leading to \eref{eq:scaling3} can be readily obtained by making a  ``minimal-disturbance hypothesis''. Assuming that adding a new training point $\vl x^*$ within the domain $\Omega_\Delta$ will only affect the dual variables of the few closest SVs, the correction of the decision function on the new SV is given by:
\begin{equation}
\label{006}
   \sum_{\abs{\vl x^\mu - \vl x^*}\leq r_c} d\alpha^\mu y^\mu \left(\frac{\abs{\vl x^\mu - \vl x^*}}{\sigma}\right)^\xi,
\end{equation}
where $d\alpha^\mu$ is the charge correction. One must have that $\sum_{\abs{\vl x^\mu - \vl x^*}\leq r_c} d\alpha^\mu y^\mu\approx -y^* \alpha^*$ to ensure that SVs further away are not affected by this perturbation. Thus $d\alpha^\mu\sim \alpha^*\sim \bar \alpha$, where the last equivalence stems from the fact that the added SV is statistically identical to any other one. Finally, requiring that the new point $\vl x^*$ must also be a SV implies that the correction represented by \eref{006} must be of order one to set $|f(\vl x^*)|=1$. Hence, we obtain the scaling relation (that implies \eref{eq:scaling3} from \eref{eq:scaling1} and \eref{eq:scaling2}):
\begin{equation}
    \boxed{\bar\alpha \left(\frac{r_c}{\sigma}\right)^\xi \sim 1.}
\end{equation}

\underline{Note 2}: The above scaling arguments may also be carried out in the intermediate regime $\delta \ll \sigma < \gamma$. In that case, the kernel \eref{eq:kertaylor} introduces a cutoff to the volume of interaction in the transverse space. In particular, the number of terms in the sum of \eref{eq:svcgradient} now goes as $(\sigma/\gamma)^{d-1} p \Delta/ \gamma$ and the average scales as $\bar\alpha \Delta \sigma^{\xi-2}$. The discussion on the fluctuations is however unaltered as $r_c \ll \sigma$ by definition. Assembling all the pieces yields the following scaling relations:
\begin{equation}\label{eq:power_laws2}
    \Delta \sim \gamma \, \left(\frac{\sigma}{\gamma}\right)^{-(d-1)\frac{d-3+\xi}{3d-3+\xi}} p^{-\frac{d-1+\xi}{3d-3+\xi}}, \quad\quad \bar\alpha \sim \left(\frac{\sigma}{\gamma}\right)^{\frac{2\xi d}{3d-3+\xi}} \, p^{\frac{2\xi}{3d-3+\xi}}, \quad\quad r_c \sim \gamma \, \left(\frac{\sigma}{\gamma}\right)^{\frac{d-3+\xi}{3d-3+\xi}} \, p^{-\frac{2}{3d-3+\xi}}
\end{equation}
and 
\begin{equation}
    \epsilon \sim \frac{\Delta}{\gamma} \sim \left(\frac{\sigma}{\gamma}\right)^{-(d-1)\frac{d-3+\xi}{3d-3+\xi}} p^{-\frac{d-1+\xi}{3d-3+\xi}}.
\end{equation}
Note that when this approach breaks down, namely when $\sigma \sim r_c$, the predictions of the vanishing bandwidth are recovered.

\subsection{Multiple interfaces}\label{sec:multiple_interfaces}
\begin{figure}[ht]
    \centerline{\includegraphics[scale=0.46]{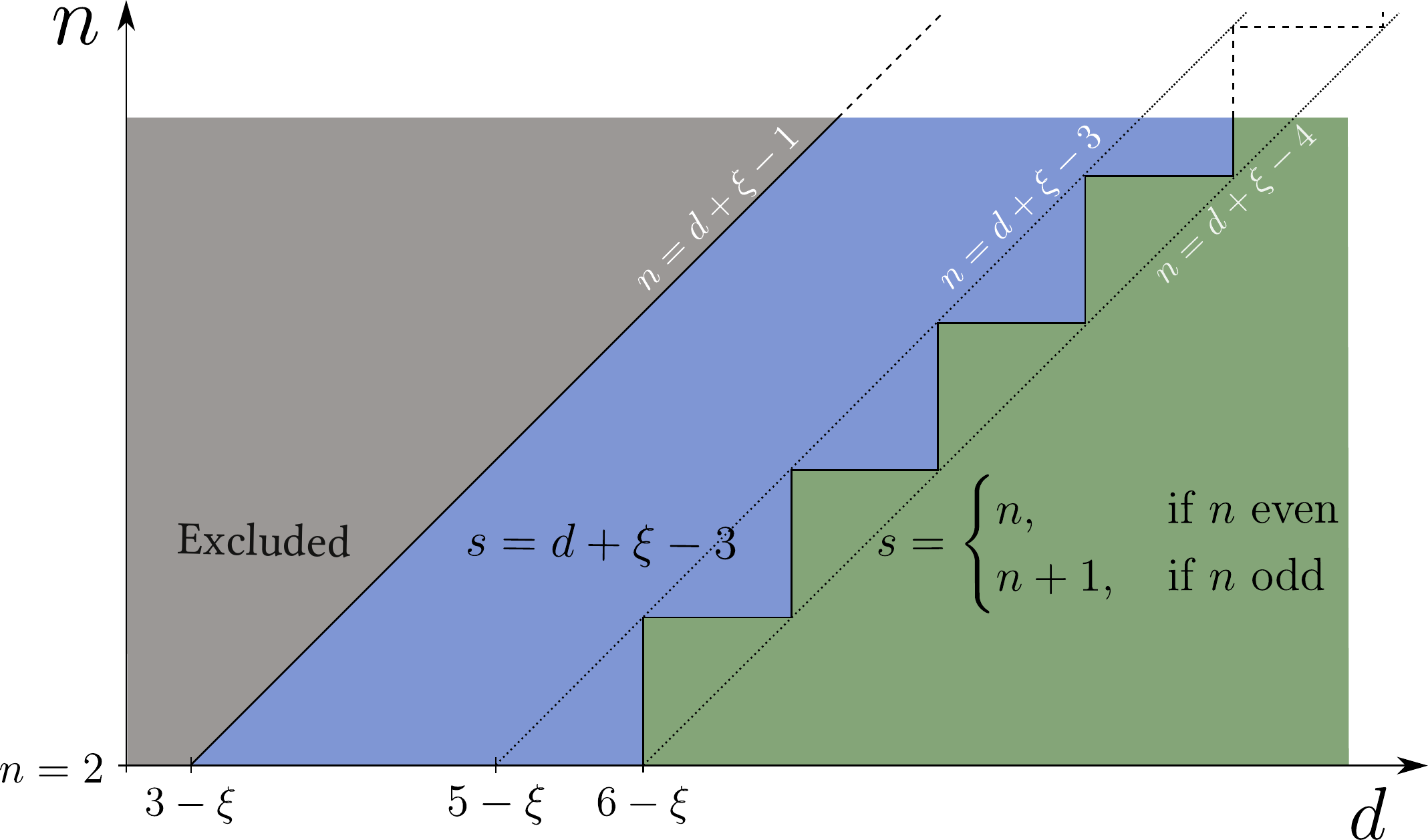}}
    \caption{\label{fig:multiple_interfaces} Sketch of the different regimes depending on the number of interfaces $n$ and the space dimension $d$. In the green region, the SVC algorithm is dominated by the large transverse contribution, $\abs{\vl x_\perp} \gg w$ ; in the blue region, it is dominated by the short transverse contributions, $\abs{\vl x_\perp} \sim w$ ; in the gray region, microscopic effect, occurring at the scale $\abs{\vl x_\perp} \sim r_c$, enter into play and have not been investigated.}
\end{figure}

The scaling analysis considered for the single interface can be directly extended to multiple interfaces. Let us consider the setup of $n$ interfaces separated by a distance $w$. Because the target function oscillates around the $n$ interfaces, its RKHS norm increases with $n$ leading to a more and more complicated task. In the limit $\Delta \ll w$, the arguments presented between \eref{eq:smoothened} and \eref{eq:scaling3} that rely on local considerations apply identically. The computation of the gradient is more subtle as the charges will in general differ in magnitude on each side of interfaces. We discuss in \aref{app:multiple_interfaces} how the resulting gradient will scale with $w$. In particular, we identify three regimes on the $(n,d)$-plane as represented on \fref{fig:multiple_interfaces}. When the dimension is large enough, in the green region, the gradient is dominated by points with large transverse distance, $\abs{\vl x_\perp} \gg w$. For smaller dimensions, the typical transverse distance decreases so that, in the blue region, the gradient is dominated by points of transverse distance $\abs{\vl x_\perp} \sim w$. For even smaller dimensions, in the gray region, our description breaks down, because the SVC function is not sufficiently smooth and microscopic effect should be accounted for. The power-laws of the three usual observables are shown to be 
\begin{equation}\label{eq:power_laws_multiple_interfaces}
    \Delta \sim \gamma \, \left(\frac{\gamma}{w}\right)^{\frac{(d-1) s}{3d-3+\xi}} p^{-\frac{d-1+\xi}{3d-3+\xi}}, \quad\quad \bar\alpha \sim \left(\frac{\sigma}{\gamma}\right)^\xi \, \left(\frac{\gamma}{w}\right)^{\frac{\xi s}{3d-3+\xi}} \, p^{\frac{2\xi}{3d-3+\xi}}, \quad\quad r_c \sim \gamma \, \left(\frac{w}{\gamma}\right)^{\frac{s}{3d-3+\xi}} \, p^{-\frac{2}{3d-3+\xi}},
\end{equation}
with
\begin{equation}
s=
\begin{cases}
n + 1, \, \mr{if} \,\, 3 \leq n \leq d + \xi - 4 \\
d + \xi -3, \, \mr{if} \,\, d + \xi - 4 \leq n \leq d + \xi - 1
\end{cases}
,\: \text{for $n$ odd},
\end{equation}
or
\begin{equation}
s=
\begin{cases}
n, \, \mr{if} \,\, 2 \leq n \leq d + \xi - 3 \\
d + \xi -3, \, \mr{if} \,\, d + \xi - 3 \leq n \leq d + \xi - 1
\end{cases}
,\: \text{for $n$ even}.
\end{equation}

The scaling in $p$ is unaltered by the presence of multiple interfaces. However, the increasing complexity of the task is reflected by the large prefactor, which requires exponentially more training points to enter the power-law decay as the width $w$ decreases. Note that for a given dimension, the task complexity, quantified by $s(d-1)/(3d-3+\xi)$, stop increasing once $n$ is large enough to enter the blue region.

\subsection{Numerical results}

In this section, we present the numerical simulations with which we verify the scalings predicted in the two previous sections. Both the single and the double-interface setups have been considered with data points sampled from an isotropic Gaussian distribution of variance $\gamma^2=1$ along each component. In the single-interface setup the hyperplane is centered at $x_1=0$, while in the double-interface setup one hyperplane is located at $x_\mr{min}=-0.3$ and the other at $x_\mr{max} \approx 1.18549$.\footnote{The value $x_\mr{max} = \sqrt{2}\,\mr{erf}^{-1}(1 + \mr{erf}(x_\mr{min})) \approx 1.18549$ is chosen in such a way that the expected number of $y=\pm1$ points is the same.} In both setups, the probability of positive and negative labels are equal. The margin-SVC algorithm is run using the class \texttt{svm.SVC} from the python library \textit{scikitlearn}, which is a soft margin algorithm. To recover the hard margin algorithm presented in \sref{sec:margin_SVC}, the regularization parameter $C$ which bounds from above the dual variables (see for example chapter 7 of \mycite{smola1998connection}) is set to $C=10^{20}$. All results presented in this section have been obtained with the Laplace kernel of bandwidth $\sigma=100 \gg \gamma$. Further results with the Matérn kernel are displayed in \aref{app:matern}.

\begin{figure}[H]
\centerline{\includegraphics[scale=.6]{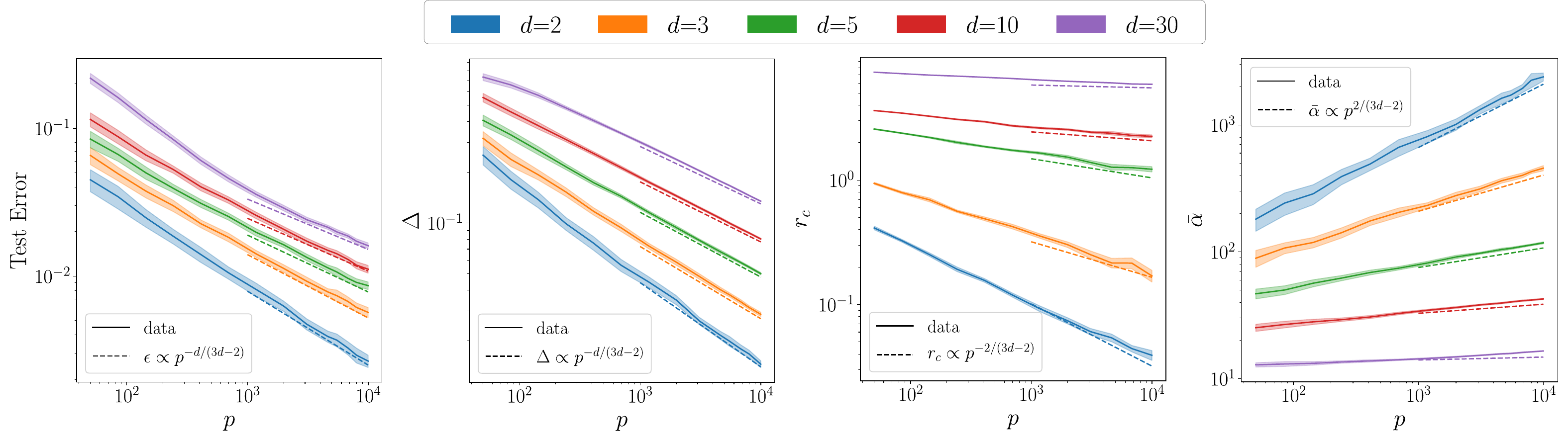}}
\caption{\label{fig:single_interface_scalings} For the \textbf{single-interface setup}, we show the dependence on the training-set size $p$ of the test error, the SV band thickness $\Delta$, the scale $r_c$ and the SV mean dual variable $\bar\alpha$ (from left to right). The points in the dataset are drawn from the standard normal distribution in dimension $d$ (see the color legend); their labels are defined according to the single-interface setup and learned with the margin-SVC algorithm  with the Laplace kernel ($\xi=1$) of bandwidth $\sigma=100$. The solid lines correspond to the average over 25 initializations, while the shaded region are the associated standard deviations. The dashed lines illustrate the power law predicted in \eref{eq:power_laws}.}
\end{figure}

\begin{figure}[H]
\centerline{\includegraphics[scale=.6]{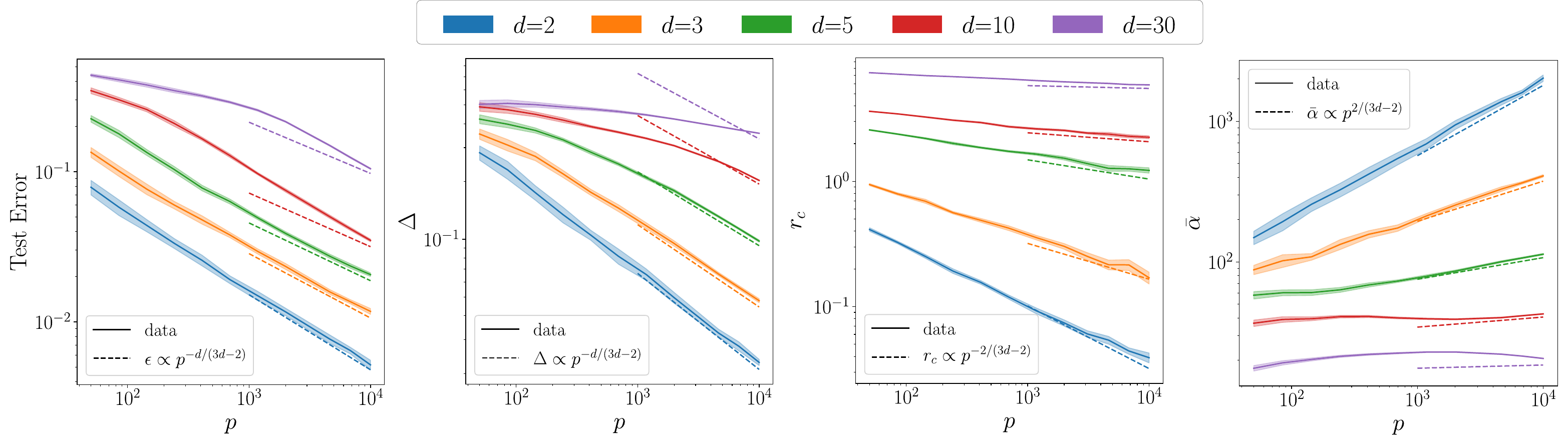}}
\caption{\label{fig:double_interface_scalings} Same plots as in \fref{fig:single_interface_scalings}, but for the \textbf{double-interface setup}: we show the dependence on the training-set size $p$ of the test error, the SV band thickness $\Delta$, the scale $r_c$ and the SV mean dual variable $\bar\alpha$ (from left to right). The points in the dataset are again drawn from the standard normal distribution in dimension $d$ (see the color legend); their labels are defined according to the double-interface setup and learned with the margin-SVC algorithm  with the Laplace kernel ($\xi=1$) of bandwidth $\sigma=100$. The solid lines correspond to the average over 25 initializations, while the shaded region are the associated standard deviations. The dashed lines illustrate the power law predicted in \eref{eq:power_laws}.}
\end{figure}

\begin{figure}[H]
\centerline{
\begin{subfigure}{.33\textwidth}
  \centering
  \includegraphics[width=1\linewidth]{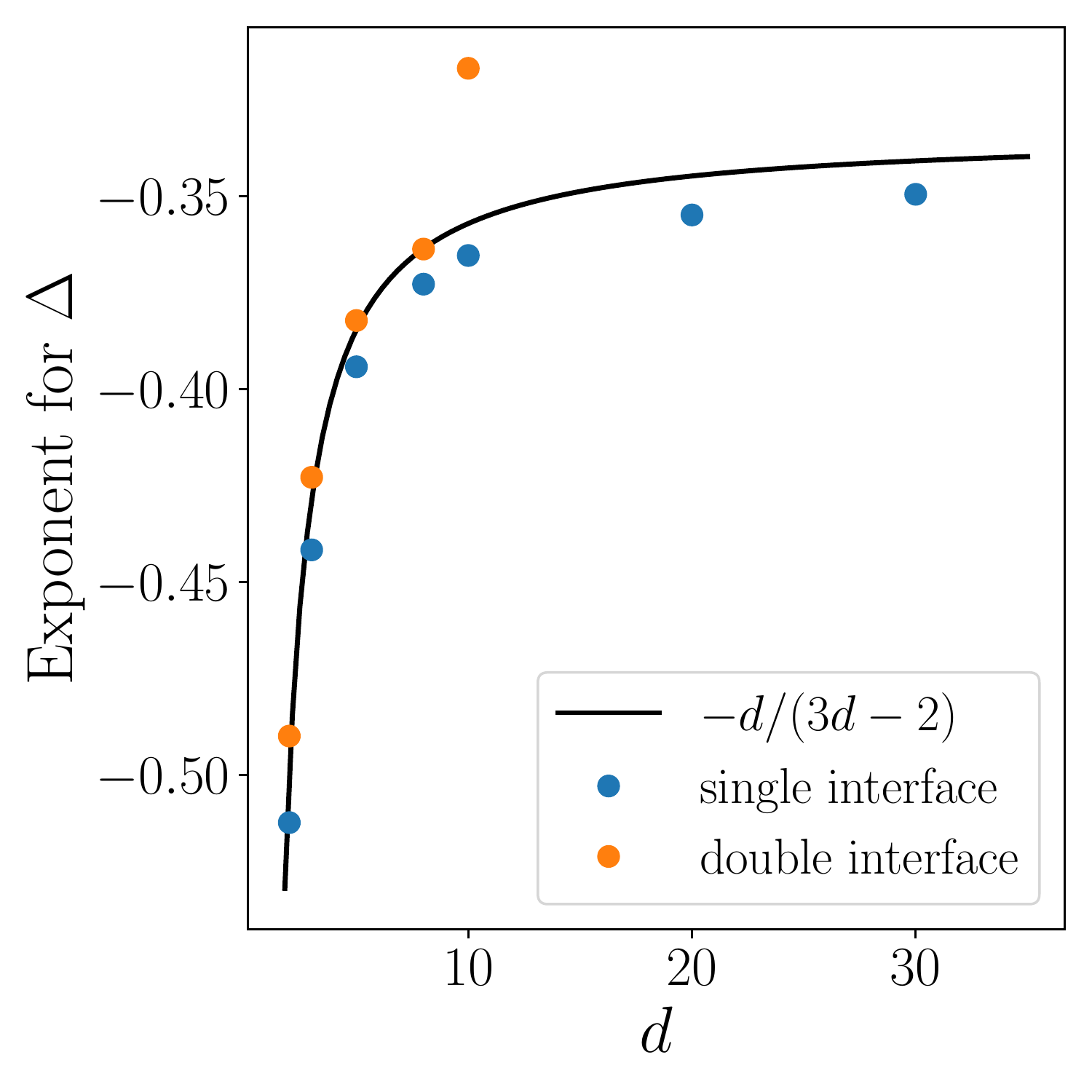}
\end{subfigure}%
\begin{subfigure}{.33\textwidth}
  \centering
  \includegraphics[width=1\linewidth]{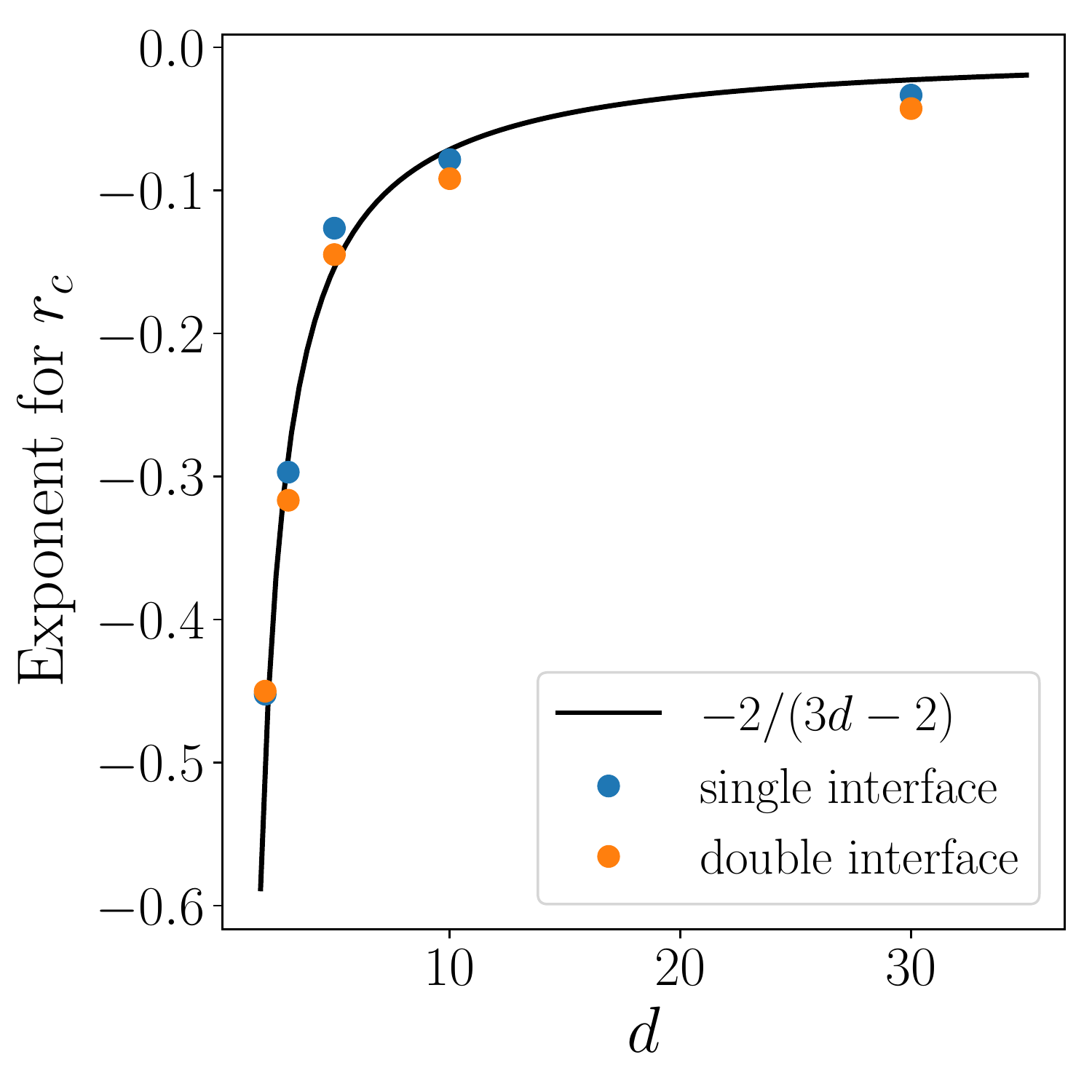}
\end{subfigure}
\begin{subfigure}{.33\textwidth}
  \centering
  \includegraphics[width=1\linewidth]{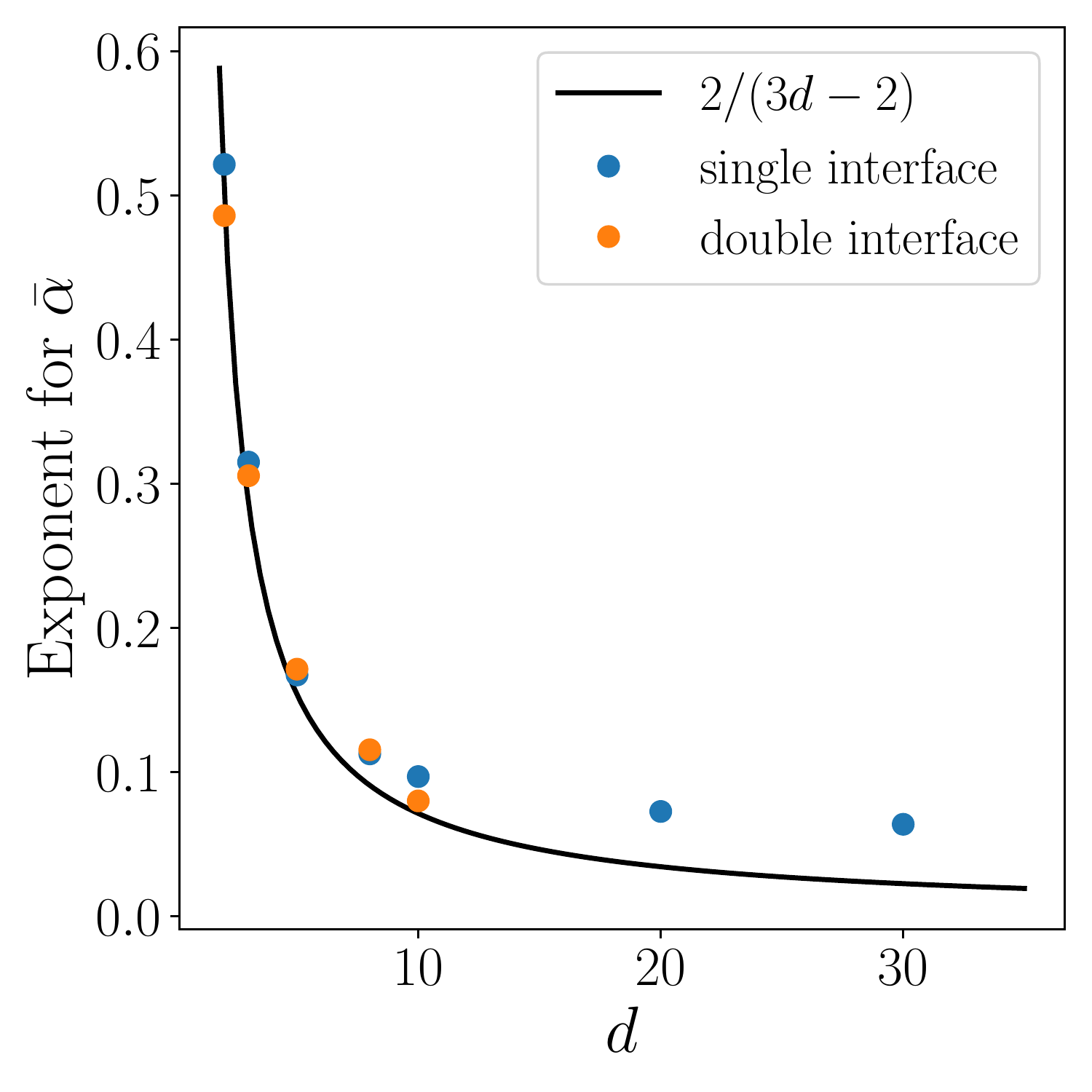}
\end{subfigure}
}
\caption{\label{fig:stripe_exponents} We extract the exponents by fitting the curves in \fref{fig:single_interface_scalings} (for the \textbf{single-interface setup}) and in \fref{fig:double_interface_scalings} (for the \textbf{double-interface setup}). We then plot the exponents for the SV band thickness $\Delta$ (left), the SV nearest-neighbor scale $r_c$ (middle) and the SV mean dual variable $\bar\alpha$ (right) against the dimension $d$ of the data. The black solid line is the prediction of \sref{sec:large_sigma}, while the dots correspond to the numerical data (blue points for the single-interface setup and orange points for the double-interface setup).}
\end{figure}

The power law predictions of \sref{sec:large_sigma} are verified in \fref{fig:single_interface_scalings} (for the single interface) and \fref{fig:double_interface_scalings} (for the double interface). The considered numerical observables are defined as follows: the test error is the fraction of mislabeled points in a test set of size $p_\mr{test}=10000$; the typical $\bar\alpha$ is the average SV dual variable; the band thickness $\Delta$ is the average distance of a SV to the closest interface; the procedure to estimate the SV nearest-neighbor scale $r_c$ is described in 
\aref{app:critical_scale}. The exponents of the power laws are extracted by fitting the numerical curves in the asymptotic regime and compared to the theoretical predictions of \sref{sec:large_sigma} in \fref{fig:stripe_exponents}. Note that in large dimensions, we observe that the system has not yet fully reached the asymptotic regime in the considered range of training-set sizes $p$. 

\begin{figure}[H]
\centerline{\includegraphics[width=.6\linewidth]{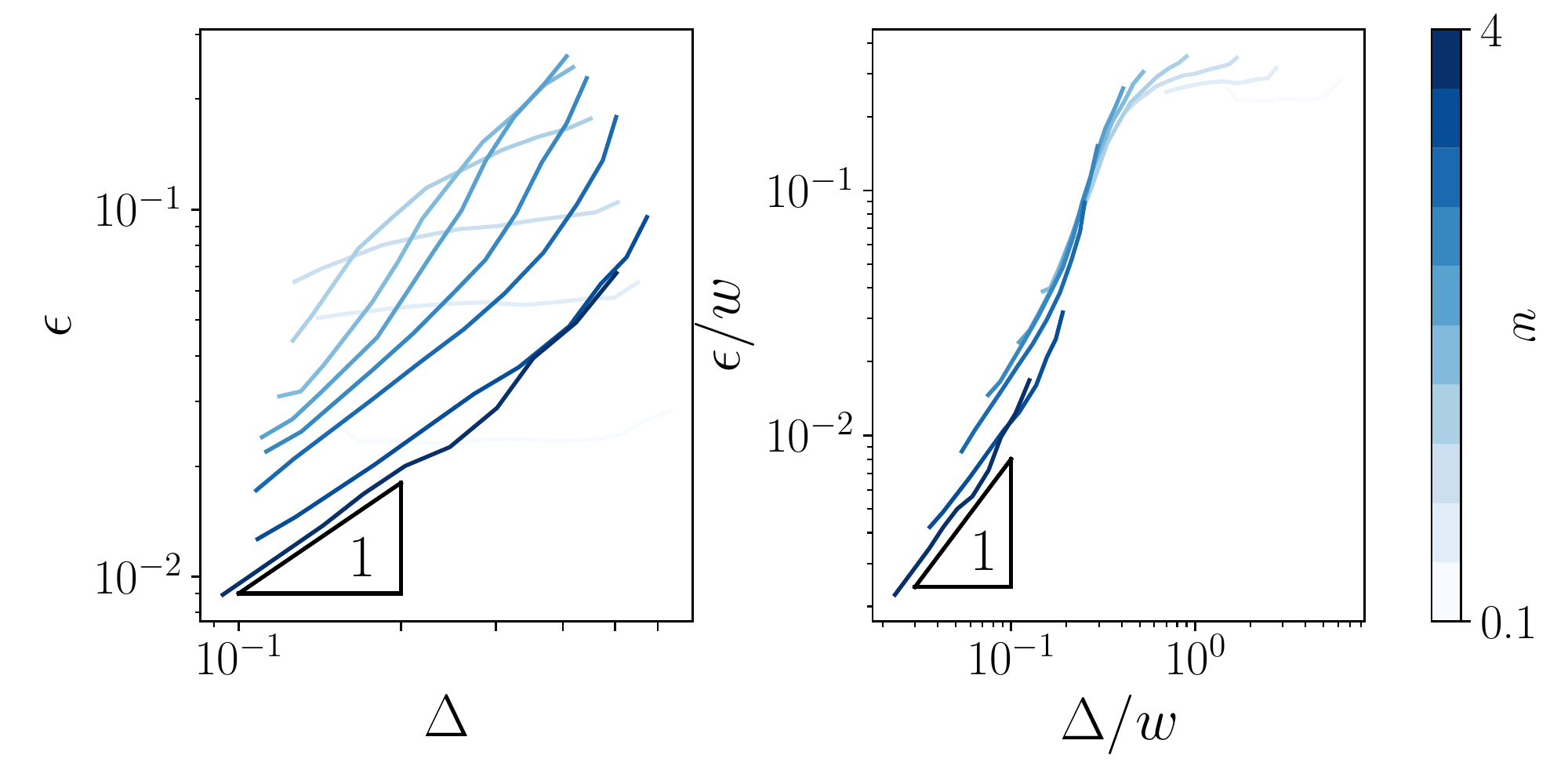}}
\caption{\label{fig:gap_collapse}\underline{Left}: Test error $\epsilon$ vs the SV band thickness $\Delta$ for multiple values of the distance between the two hyperplanes $w$ for the \textbf{double-interface setup} in $d=5$. The left interface is located at $x_\mr{min}=-1$ and the right interface at $x_\mr{max}=x_\mr{min} + w$. \underline{Right}: The left plot is rescaled by $w^{-1}$ both horizontally and vertically. The inset triangles indicate a slope of one in log-log scale.}
\end{figure}

We also observe that in the double-interface setup, the system only enters the scaling regime when $\Delta$ becomes small enough compared to the distance $w$ between the two hyperplanes, as discussed in \sref{sec:multiple_interfaces}. The crossover from the interfering-interfaces regime to the asymptotic regime is illustrated in \fref{fig:gap_collapse}. The test error vs $\Delta$ displayed on the left figure for multiple values of $w$ confirms that $\epsilon \sim \Delta$, when $\Delta \ll w$, as expected from the discussion of \sref{sec:multiple_interfaces}. We show on the right figure that the transition to the asymptotic regime occurs when $\Delta \sim w$ by rescaling the horizontal axis: $\Delta \to \Delta / w$. Because $\epsilon \sim \Delta$ in the asymptotic regime, it is necessary to also rescale the vertical axis for the curves to collapse, namely $\epsilon \to \epsilon / w$.

\section{Spherical model}\label{sec:sphere}

\begin{figure}[H]
\centerline{\includegraphics[scale=.6]{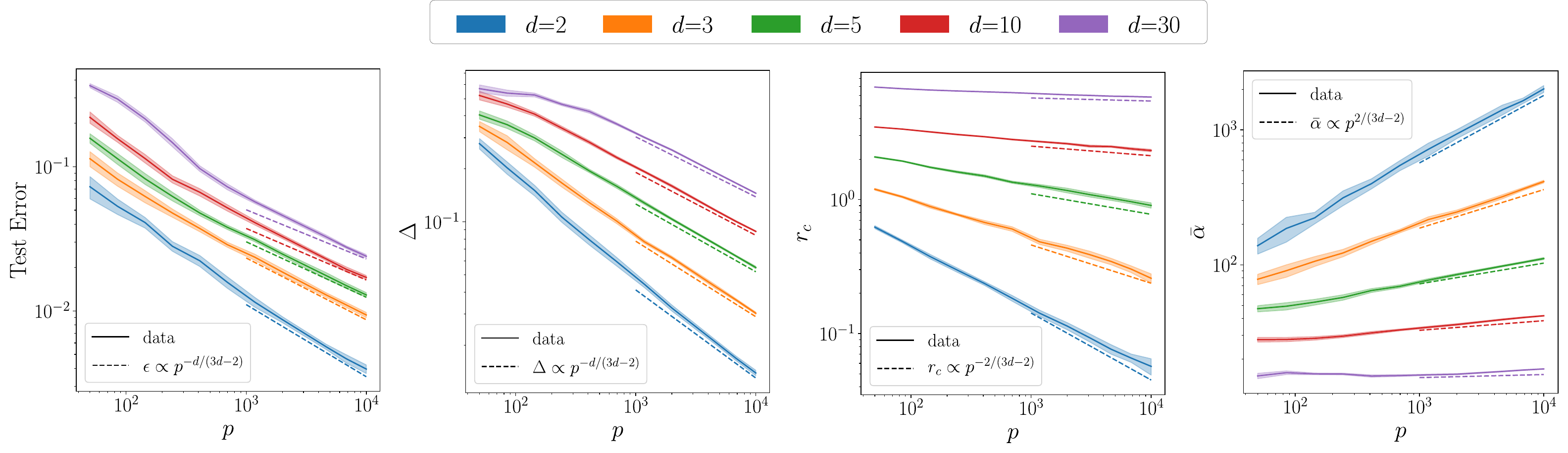}}
\caption{\label{fig:sphere_scalings} For the \textbf{spherical setup}, we show the dependence on the training-set size $p$ of the test error, the SV band thickness $\Delta$, the scale $r_c$ and the SV mean dual variable $\bar\alpha$ (from left to right). The points in the dataset are drawn from the standard normal distribution in dimension $d$ (see the color legend); their labels are defined according to the spherical setup of radius $R=\sqrt{d}$ and learned with the margin-SVC algorithm  with the Laplace kernel ($\xi=1$) of bandwidth $\sigma=100$. The solid lines correspond to the average over 25 initializations, while the shaded region are the associated standard deviations. The dashed lines illustrate the power law predicted in \eref{eq:power_laws}.}
\end{figure}

\begin{figure}[H]
\centerline{
\begin{subfigure}{.33\textwidth}
  \centering
  \includegraphics[width=1\linewidth]{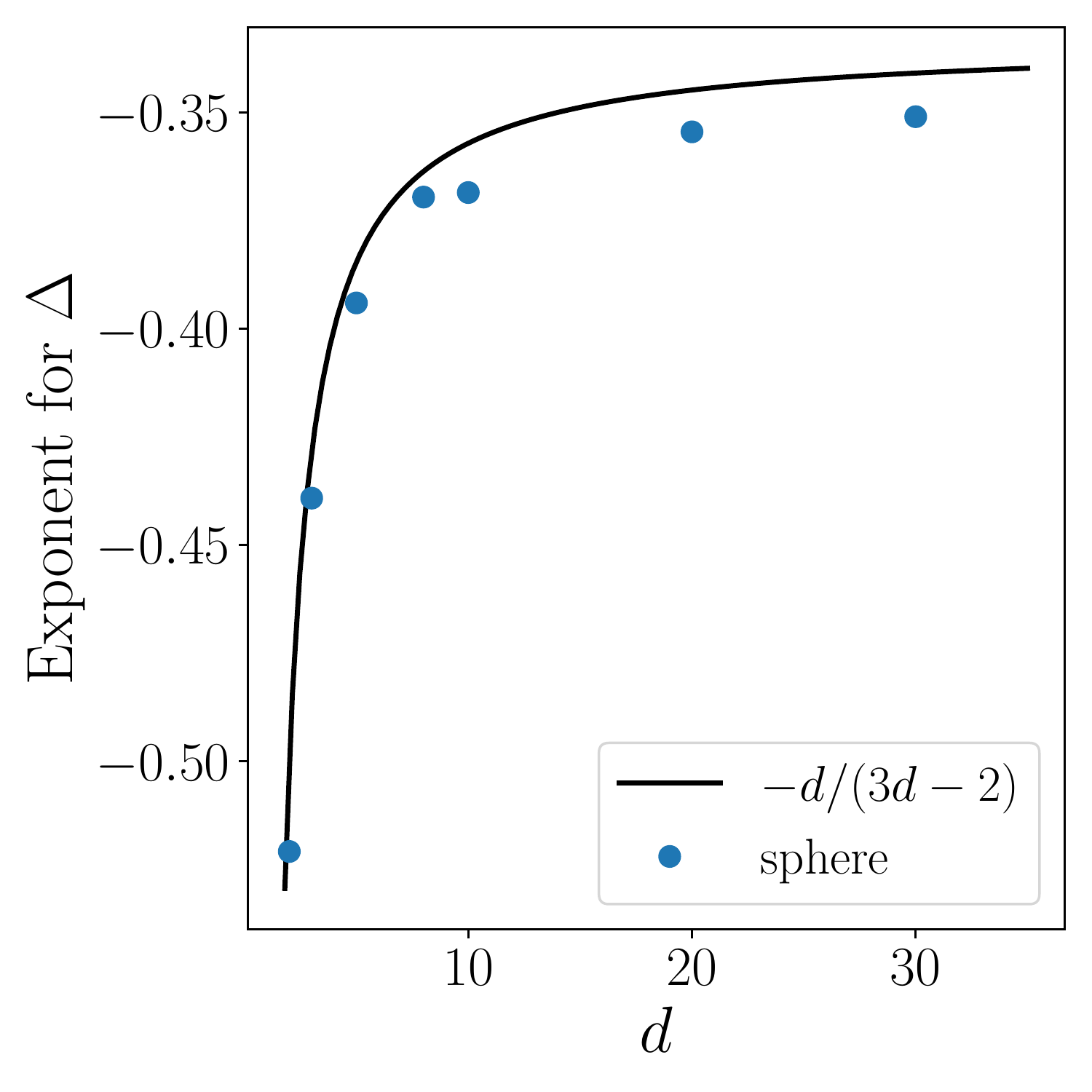}
\end{subfigure}%
\begin{subfigure}{.33\textwidth}
  \centering
  \includegraphics[width=1\linewidth]{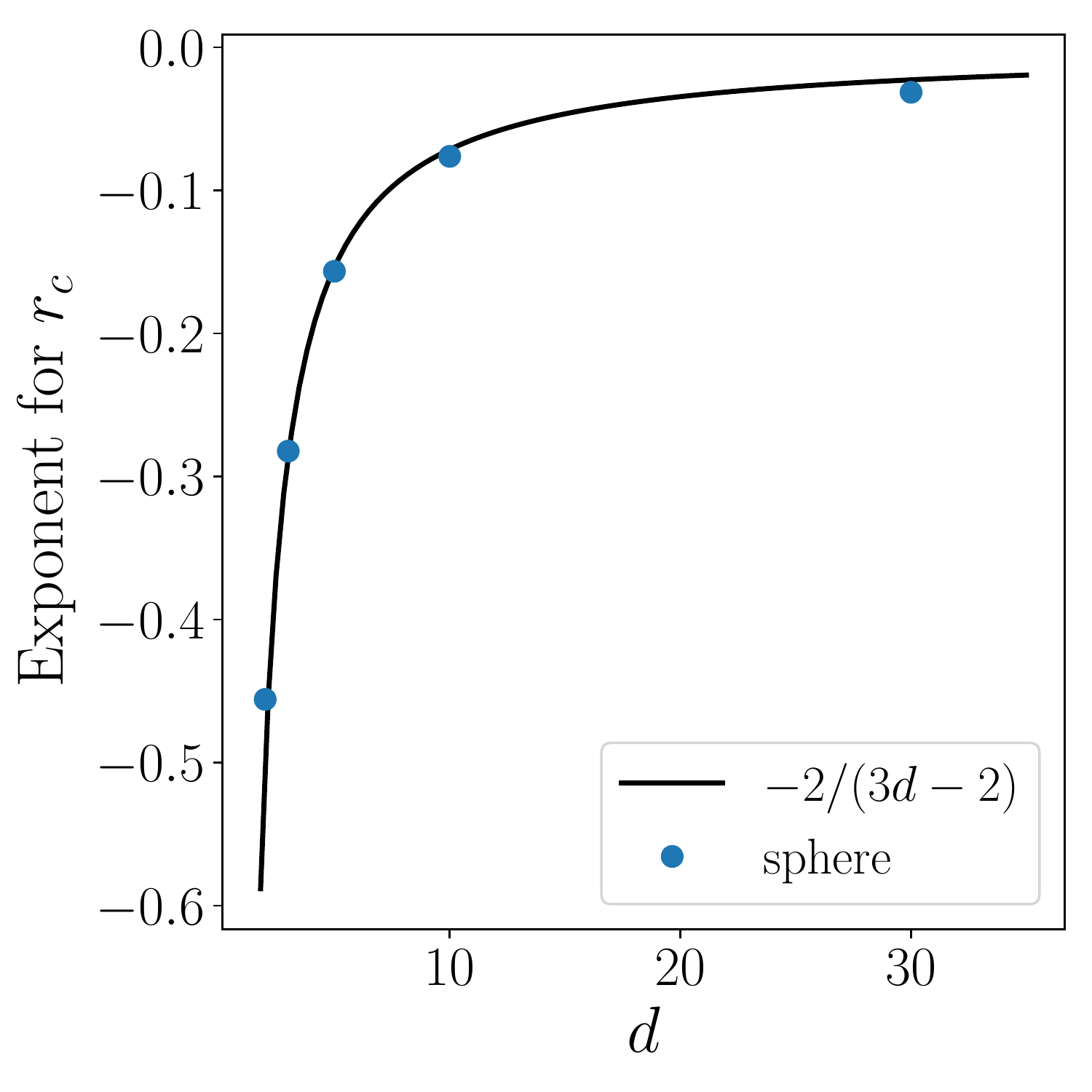}
\end{subfigure}
\begin{subfigure}{.33\textwidth}
  \centering
  \includegraphics[width=1\linewidth]{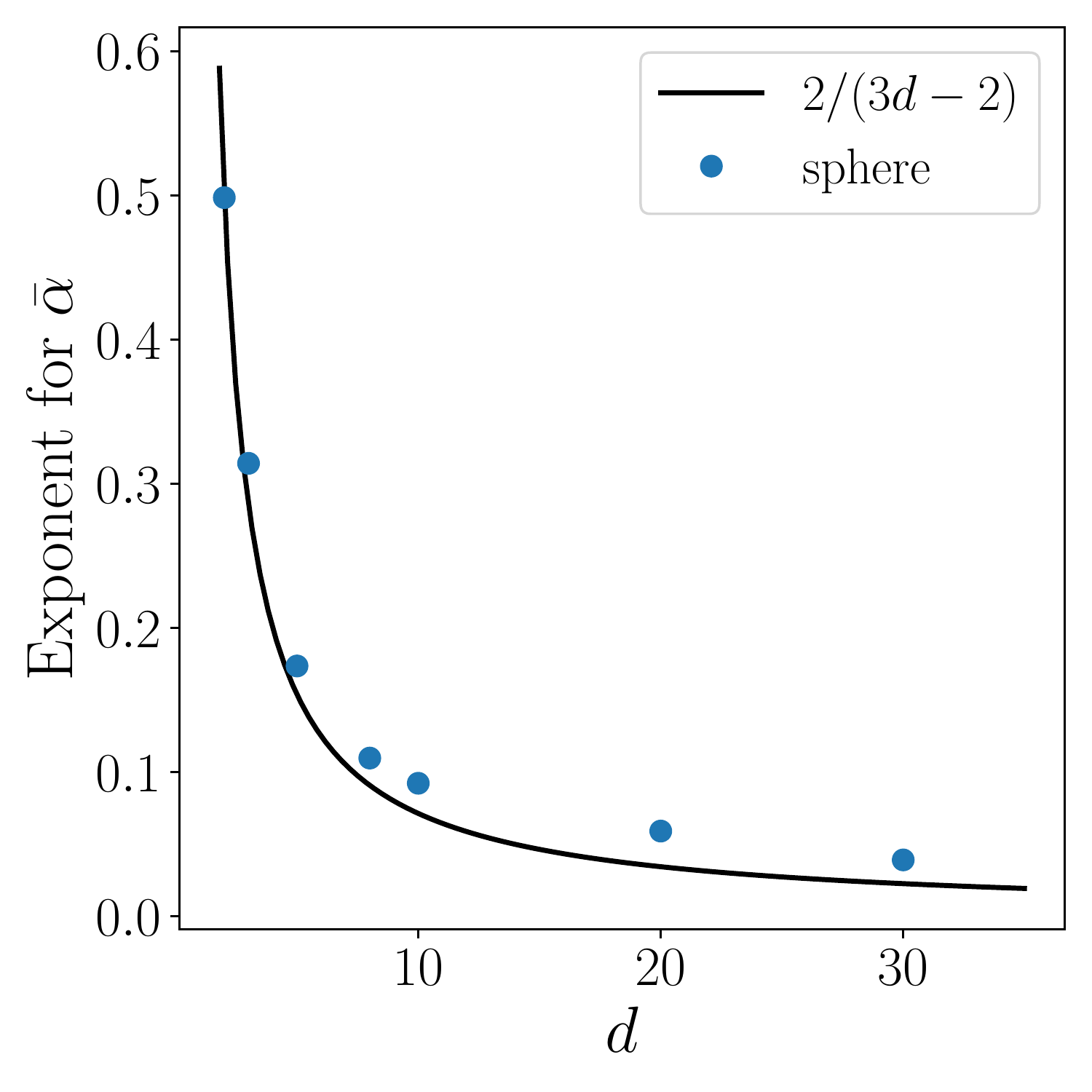}
\end{subfigure}
}
\caption{\label{fig:sphere_exponents} We extract the exponents by fitting the curves in \fref{fig:sphere_scalings} for the \textbf{spherical setup}. We then plot the exponents for the SV band thickness $\Delta$ (left), the SV nearest-neighbor scale $r_c$ (middle) and the SV mean dual variable $\bar\alpha$ (right) against the dimension $d$ of the data. The black solid line is the prediction of section \sref{sec:large_sigma}, while the dots correspond to the numerical data (blue points for the single-interface setup and orange points for the double-interface setup).}
\end{figure}

We  consider a spherical interface separating $y=+1$ points outside a sphere of radius $R$ from $y=-1$ points inside. The relevant direction is therefore $x_\parallel = \abs{\vl x}$, and the label is given by $y(\vl x) = \mr{sign}(\abs{\vl x} - R)$. 
\begin{figure}[ht]
    \centering
    \includegraphics[scale=0.5]{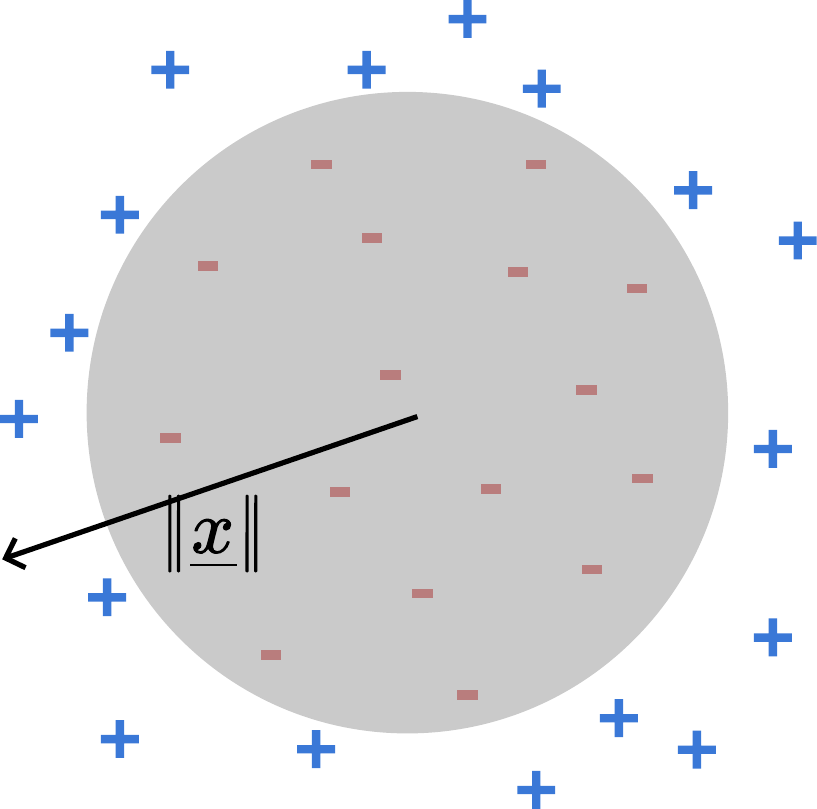}
    \caption{\label{fig:kernelsphereinterface} Nonlinear decision boundary for the spherical setup. The label function is $y=-1$ inside the hypersphere and $y=+1$ outside. Note that the label only depends on the norm of the data, $\abs{\vl x}$.}
\end{figure}
We still assume that the SV are distributed along the interface, thus forming a shell of radius $R$ and thickness $\Delta$. Once again, previous  arguments presented between \eref{eq:smoothened} and \eref{eq:scaling3} that rely on local considerations apply identically. Furthermore, we compute in  \aref{app:spherescalings} the gradient $\partial f/\partial x_\parallel $ and find again the same asymptotic result as for planar interface specified in Eq.\ref{eq:svcgradient}. Thus our predictions for the spherical model are identical to the ones for the stripe model. We test these results numerically for a sphere of radius $R=\sqrt{d}$\footnote{It guarantees that the fraction of positive and negative labels remain finite. In particular, in the limit $d\to\infty$, this fraction goes to $1/2$.} with a Laplace kernel of variance $\sigma=100$. The results displayed on \fref{fig:sphere_scalings} and \fref{fig:sphere_exponents} confirm our analysis. 

\section{Improving kernel performance by compressing invariants}\label{sec:stretching}

In this section, we investigate how compressing the data along the irrelevant directions $\vl x_\perp$ affects the performance of kernel classification. This analysis is of particular interest for neural networks, where it is now argued (see for instance \mycite{mallat2016understanding}) that a progressive capability to compress invariants in the data is built up moving through the layers of deep networks.

\subsection{Stripe model}

\begin{figure}[ht]
\centering
\includegraphics[width=1\linewidth]{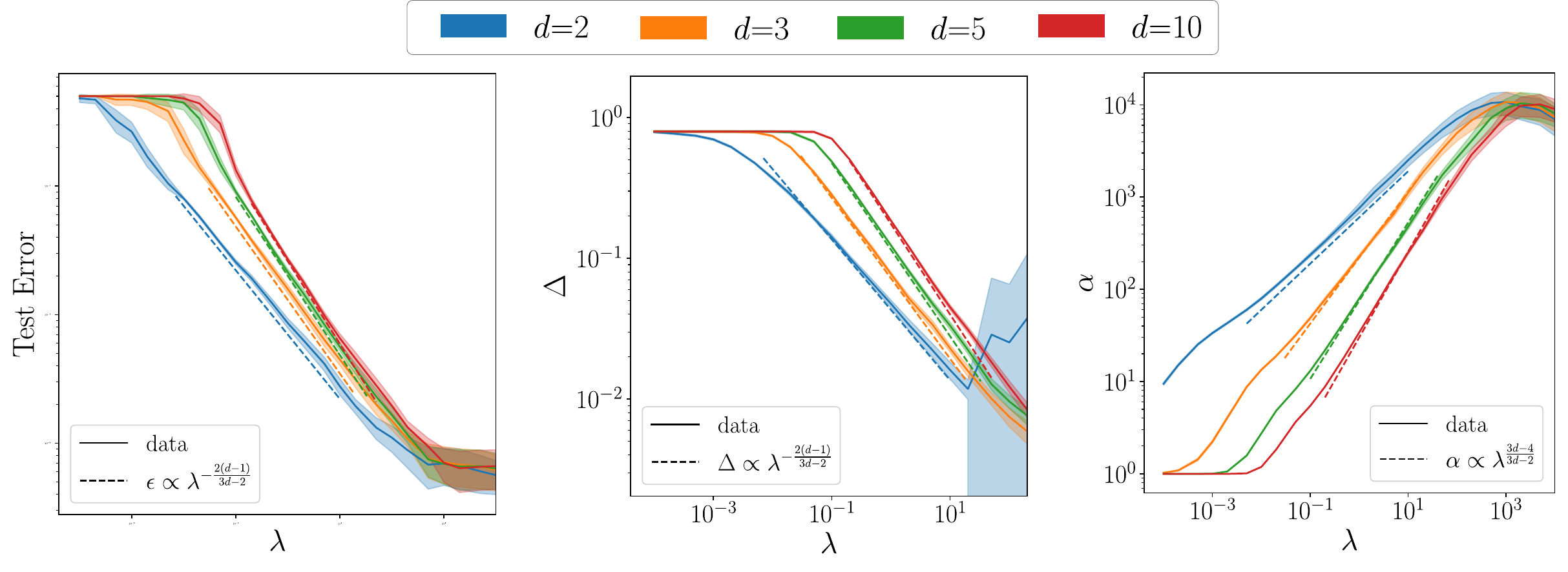}
\caption{Dependence on the amplification factor $\lambda$ of the test error (left), the SV band thickness $\Delta$ (middle) and the SV mean dual variable $\bar\alpha$ (right) for the \textbf{single-interface setup} with $p=1000$ in different dimensions (see the color legend). The SVC algorithm is run with the Laplace kernel ($\xi=1$) of bandwidth $\sigma=100 \gg \delta$. The solid lines correspond to the average over 20 initializations and the shaded region are the associated standard deviations. The dashed lines illustrate the power law predictions of \eref{eq:stretching}.}
\label{fig:SVCcompression}
\end{figure}

We consider the stripe model of \sref{sec:svc} with one additional parameter: the amplification factor $\lambda$. If the original distribution was characterized by the scales $\gamma_1,\dots,\gamma_d$ along each space direction, we now apply a contraction in the transverse space: $\gamma_i \to \gamma_i/\lambda$ for $i=2,\dots,d$. 
Following the same reasoning as in \sref{sec:large_sigma}, we can track the effect of the additional amplification parameter. It is not sufficient to merely rescale $\gamma$, since the compression is not isotropic. Nevertheless, it is easy to see that the first scaling becomes 
\begin{equation}\label{eq:scaling1_compression}
    \lambda^{d-1} \,r_c^{d-1} \, \Delta \, p \sim \gamma^d,
\end{equation}
since the density of points inside the SV band is now $\sim p \lambda^{d-1} / \gamma^d $. Then, for the second scaling relation, we need to rescale the gradient $\partial_{x_1} f$ defined in \eref{eq:svcgradient}. The amplification factors only alters the transverse space: when approximating the average by an integral, the boundaries are rescaled to $\gamma/\lambda$ in each transverse direction. The second scaling is thus
\begin{equation}\label{eq:scaling2_compression}
    \lambda^{2-\xi} \, p \, \bar\alpha \, \left(\frac{\Delta}{\gamma}\right)^3 \, \left(\frac{\gamma}{\sigma}\right)^\xi \sim 1.
\end{equation}
Finally, when imposing that the fluctuations between $f$ and its smoothed version $\tl f$ are of order one, one only needs to update the volume of the transverse space in \eref{eq:fluctuation_over_space}: $\gamma^{d-1} \to (\gamma/\lambda)^{d-1}$, which leads to the last scaling,
\begin{equation}\label{eq:scaling3_compression}
    \lambda^{d-1} \, \bar\alpha^2 \, p \, \frac{\Delta}{\gamma} \, \left(\frac{r_c}{\gamma}\right)^{d-1+2\xi} \sim \left(\frac{\sigma}{\gamma}\right)^{2\xi}.
\end{equation}
Assembling all the scaling relations yields:
\begin{equation}\label{eq:stretching}
    \epsilon \sim \Delta \sim \gamma \, \lambda^{-\frac{2(d-1)}{3d-3+\xi}} \,  p^{-\frac{d-1+\xi}{3d-3+\xi}}, \quad\quad \bar\alpha \sim \left(\frac{\sigma}{\gamma}\right)^\xi \, \lambda^{\xi\frac{3d-5+\xi}{3d-3+\xi}} \, p^{\frac{2\xi}{3d-3+\xi}}, \quad\quad r_c \sim \gamma \, \lambda^{-\frac{3d-5+\xi}{3d-3+\xi}} \, p^{-\frac{2}{3d-3+\xi}}.
\end{equation}
These power laws are assessed numerically for the Laplace kernel $(\xi=1)$ of variance $\sigma=100$ and a training set of size $p=1000$ generated from the Gaussian distribution of variance $\gamma^2=1$. Varying the amplification factor over eight orders of magnitude (see \fref{fig:SVCcompression}), our predictions hold in a broad  range of $\lambda$ but break down at large and small values, as we now explain. 

In the limit $\lambda\to0$, the relevant direction $x_1$ is negligibly small compared to the other directions, information is thus suppressed and points are classified at random: the test error goes to $\nicefrac12$. Furthermore, all training points must be SVs, and indeed $\Delta\to\<|x|\>_{x\sim\mathcal{N}(0, 1)} = \sqrt{2/\pi}$ (which is the average distance from any point in the dataset to the interface) and $\bar\alpha\to1$. 

In the opposite limit $\lambda\to\infty$ the setup lives in dimension one (seeing only $x_1$) and all curves converge independently of the space dimension $d$. These relations allow us to identify a critical scale $\lambda_c$ at which the multidimensional system reduces effectively to a one dimensional system. It occurs when the test error of the compressed multidimensional kernel is equal to the test error of the kernel that only sees the component $x_1$. Using our scalings, we find
\begin{equation}
    \lambda^{-\frac{2(d-1)}{3d-3+\xi}} \,  p^{-\frac{d-1+\xi}{3d-3+\xi}}\sim p^{-1} \hspace{15pt} \implies \hspace{15pt} \lambda_c \sim p.
\end{equation}

\subsection{Cylinder model}
We now consider a cylinder model in $d = d_\parallel + d_\perp$ dimension. A point $\vl x = (\vl x_\parallel, \vl x_\perp)\in \mathbb{R}^d$ (with $\vl x_\parallel \in \mathbb{R}^{d_\parallel}$ and $\vl x_\perp \in \mathbb{R}^{d_\perp}$) has a positive label if $\abs{\vl x_\parallel} > R \sim \gamma$ and a negative label otherwise. Such a model is also characterized by the asymptotic scalings in $p$ specified in \eref{eq:svcgradient}.

As in the previous section, we compress the perpendicular directions by the amplification factor $\lambda$: $\vl x_\perp \to \vl x_\perp / \lambda$. The derivations of the scaling relations \eref{eq:scaling1_compression} and \eref{eq:scaling3_compression} hold equally. However, the scaling relation \eref{eq:scaling2_compression} is now independent on the amplification factor: the characteristic size of the transverse space occurring in the gradient integral \eref{eq:svcgradient} remains of the order of the system size $\gamma$. Assembling the different scalings yields:
\begin{equation}\label{eq:stretching_cylinder}
    \epsilon \sim \Delta \sim \gamma \, \lambda^{-\frac{\xi d_\perp}{3d-3+\xi}} \,  p^{-\frac{d-1+\xi}{3d-3+\xi}}, \quad\quad \bar\alpha \sim \left(\frac{\sigma}{\gamma}\right)^\xi \, \lambda^{\frac{3 \xi d_\perp}{3d-3+\xi}} \, p^{\frac{2\xi}{3d-3+\xi}}, \quad\quad r_c \sim \gamma \, \lambda^{-\frac{3 d_\perp}{3d-3+\xi}} \, p^{-\frac{2}{3d-3+\xi}}.
\end{equation}

\section{Conclusion}

We have studied the learning curve exponent $\beta$ of isotropic kernel in the presence of invariants, improving on worst case bounds previously obtained in the literature. For regression on Gaussian fields, we find that invariants do not increase $\beta$ that behaves as $\sim d^{-1}$ in large dimension: methods based on isotropic kernels suffer from the curse of dimensionality,  as already argued in (\mycite{bach2017breaking}). Our analysis also suggests a simple estimate \ref{int} for the performance of regression beyond the Gaussian fields considered here.
For a binary classification and simple models of invariants we find the opposite result. For a planar interface separating labels,  $\beta\geq 1/3$ for all dimensions, improving on previous bounds. 

Note that the striking difference between classification and regression does not stem from the distinct models considered in each case. Indeed, following \eref{int} we expect that performing mean-square ridgeless regression on the stripe model leads to the curse of dimensionality with $\beta=1/d$, as we have checked on a few examples (data not shown). In the classification problem instead, due to the fact that only a tiny band of data are support vectors, the output function ends up being much smoother (i.e. with more rapidly decaying Fourier components) than a step function, leading to better performance.

This success of classification holds when several interfaces are present, or in the spherical case where the interface continuously bends. Thus, isotropic kernels can beat the curse of dimensionality even for non-planar  boundaries between labels. For which class of boundaries is  this result true? 
The geometry of the spatial distribution of support vectors suggests an intuitive answer. 
The curse of dimensionality is beaten because a very narrow (i.e. rapidly decaying with $p$)  layer of width $\Delta$ is sufficient to fit all data, despite the fact that the distance between support vectors $r_c$ is much larger (and indeed subjected to the curse of dimensionality). Thus if the boundary displays significant variations below the scale $r_c$, it presumably cannot be detected by isotropic kernels. In that view, beating the curse of dimensionality is only possible if the boundary is more and more regular as the dimension increases. This geometrical view is consistent with the more abstract kernel literature in which the curse is lifted if labels correspond to the sign of a  regular function (in the sense of belonging to the RKHS of the kernel) \cite{bartlett2002rademacher}. Empirically, sufficient regularity may be achieved in practical settings at least along some invariants, such as completely uninformative pixels near the boundary of  images. Under which conditions  other invariants, e.g. related to translation, can be exploited by isotropic kernels remains to be understood. 

Note added: In \cite{paccolat2020compressing}, these results were extended beyond kernels, to the case of a wide one-hidden layer net. In the lazy training regime, results are identical to those presented here, but more favorable exponents $\beta$ are  found in the feature learning regime.

\subsection*{Acknowledgments}
We acknowledge L. Chizat, M. Geiger, P. Loucas, L. Petrini, C. Pehlevan for  discussions and L. Chizat for pointing out several important references.
This work was partially supported by the grant from the Simons Foundation (\#454953 Matthieu Wyart). M.W. thanks the Swiss National Science Foundation for support under Grant No.~200021-165509.

\clearpage
\bibliographystyle{unsrt}
\bibliography{main}{}

\clearpage
\appendix

\section{Kernel regression with invariant dimensions} \label{app:formalregrTSthm}

\vspace{3em}\textbf{Theorem} Let $K_T(\vl x)$ and $K_S(\vl x)$ be two translation-invariant kernels (called the \emph{Teacher} and \emph{Student} respectively) defined on $\mc V_d \equiv \mb R^d$, and let $\tl K_T(\vl w)$ and $\tl K_S(\vl w)$ be their Fourier transforms in $\mc V_d$. Assume that
\begin{itemize}
    \item $K_T(\vl x), K_S(\vl x)$ are continuous everywhere and differentiable everywhere except at the origin $\vl x=0$;
    \item $K_T(\vl x)$ and $K_S(\vl x)$ are positive definite and isotropic, that is, they only depend on $\abs{\vl x}$;
    \item $K_T(\vl x)$ and $K_S(\vl x)$ have a cusp at the origin and their $d$-dimensional Fourier transform decays at high frequencies with dimensional-dependent exponents $\alpha_T(d_\parallel)$ and $\alpha_S(d)$, respectively (we will evaluate them at $d_\parallel$ for the Teacher and at $d$ for the Student);
    \item $\lim_{\vl x\to0} K_T(0) < \infty$ and $\lim_{\vl x\to0} K_S(0) < \infty$;
    \item $\lim_{\vl w\to0} \tl K_T(\vl w) < \infty$ and $\lim_{\vl w\to0} \tl K_S(\vl w) < \infty$.
\end{itemize}
Assume furthermore that the Teacher kernel lives in a reduced space of dimension $d_\parallel\leq d$, in the sense that
\begin{itemize}
    \item $K_T(\vl x) \equiv K_T(x_1, \dots, x_d) = K_T(x_1,\dots,x_{d_\parallel}) \equiv K_T\left(\abs{\vl x_\parallel}\right)$ (where we have defined $\vl x_\parallel \equiv (x_1,\dots,x_{d_\parallel})^t$).
\end{itemize}
We use the Teacher kernel to sample a Gaussian random field $Z_T(\vl x) \sim \mc N(0,K_T)$ at points that lie on a $d$-dimensional regular lattice in $\mc V_d$, with fixed spacing $\delta$, and we use the Student kernel to infer $\hat Z_S(\vl x)$ at a new point $\vl x\in\mc V_d$ via regression, and performance is then evaluated by computing the expected mean-squared error on points independent from those used for training.
Then, as $\delta\to0$,
\begin{equation}
    \mb E\,\mr{MSE} \sim \delta^{\beta d} \quad \mr{with} \quad \beta=\frac1d\min(\alpha_T(d_\parallel)-d_\parallel, 2\alpha_S).
\end{equation}

\textbf{Proof.}

\emph{(i) Set-up.}

We first consider a finite number of points $p$ in a box $\mc V_d = [-\nicefrac{L}2,\nicefrac{L}2]^d$ and then take the limit $p,L\to\infty$, keeping the spacing $\delta = Lp^{-\nicefrac{1}{d}}$ fixed. Regression is done by minimizing the mean-squared error on the $p$ points:
\begin{equation}
    \sum_{\mu=1}^p\left[Z_T(\vl x_\mu) - \hat Z_S(\vl x_\mu)\right]^2,
\end{equation}
and the generalization error is defined as
\begin{equation}
    \mb E\,\mr{MSE} = L^{-d} \mb E \int_{\mc V_d}\mr{d}^d\vl x\,\left[Z_T(\vl x) - \hat Z_S(\vl x)\right]^2.
\end{equation}
(The expectation value is taken with respect to the Teacher random process).

Given a function $F(\vl x)$ on the $d$-dimensional box $\mc V_d = [-\nicefrac{L}{2},\nicefrac{L}{2}]^d$, we denote its Fourier transform (series) and antitransform by
\begin{align}
&\tl F(\vl w) \equiv \mc{F}_d\left[F(\vl x)\right](\vl w) = L^{-\nicefrac{d}2} \int_{\mc V_d} \mr{d}^d\vl x\, e^{-i\vl w\cdot\vl x} F(\vl x), \quad \mr{where}\ \vl w \in \mb L_d\equiv\frac{2\pi}L \mb{Z}^d,\\
&F(\vl x) \equiv \mc{F}^{-1}_d\left[\tl F(\vl w)\right](\vl x) = L^{-\nicefrac{d}2} \sum_{\vl w\in\mb{L}} e^{i\vl w\cdot\vl x} \tl F(\vl w).
\end{align}

Given the structure of the Teacher kernel we can write
\begin{multline}
    \tl K_T(\vl w) = L^{-\nicefrac{d_\parallel}2} \int_{[-\nicefrac{L}{2},\nicefrac{L}{2}]^{d_\parallel}} \mr{d}^{d_\parallel}\vl x_\parallel e^{-i\vl w_\parallel\cdot\vl x_\parallel} K_T\left(\abs{\vl x_\parallel}\right) \cdot L^{-\nicefrac{d_\perp}{2}} \int_{[-\nicefrac{L}{2},\nicefrac{L}{2}]^{d_\perp}} \mr{d}^{d_\perp}\vl x_\perp e^{-i\vl w_\perp\cdot\vl x_\perp} = \\
    = \mc{F}_{d_\parallel}\left[K_T\left(\abs{\vl x_\parallel}\right)\right](\vl w_\parallel) \cdot L^{\nicefrac{d_\perp}{2}} \delta_{\vl w_\perp}.\label{eq:ftransparall}
\end{multline}
This formula states that the Fourier transform of the Teacher kernel has frequencies that also live in the corresponding $d_\parallel$-dimensional subspace in the frequency domain. The term $\delta_{\vl w_\perp}$ is a discrete delta (not a Dirac delta): this will be important later because it implies that it is scale invariant: $\delta_{a\vl w_\perp} = \delta_{\vl w_\perp}$. The first term, that is the Fourier transform of the Teacher kernel restricted to the $d_\parallel$-dimensional space, decays at large frequencies with an exponent $\alpha_T(d_\parallel)$ that depends on the intrinsic dimension $d_\parallel$:
\begin{equation}
    \tl K_T^\parallel(\vl w_\parallel) \equiv \mc{F}_{d_\parallel}\left[K_T\left(\abs{\vl x_\parallel}\right)\right](\vl w_\parallel) = c_T(d_\parallel) \abs{\vl w_\parallel}^{-\alpha_T(d_\parallel)} + o\left(\abs{\vl w_\parallel}^{-\alpha_T(d_\parallel)}\right).
\end{equation}

\emph{(ii) Regression.}

The solution to the regression problem can be computed in closed form:
\begin{equation}
\hat Z_S(\vl x) = \vl k_S(\vl x) \cdot \mb K_S^{-1} \vl Z_T,
\end{equation}
where where $\vl Z_T = \left(Z_T(\vl x_\mu)\right)_{\mu=1}^p$ are the training data (the points $\vl x_\mu$ lie on the regular lattice), $\vl k_S(\vl x) = \left(K_S(\vl x_\mu, \vl x)\right)_{\mu=1}^p$ and $\mb K_S = \left(K_S(\vl x_\mu, \vl x_\nu)\right)_{\mu,\nu=1}^p$ is the Gram matrix, that is invertible since the kernel $K_S$ is assumed to be positive definite. This formula can be written in Fourier space as
\begin{equation}
\tl Z_S(\vl w) = \tl Z^\star(\vl w) \frac{\tl K_S(\vl w)}{\tl {K_S}^\star(\vl w)},\label{eq:zpost}
\end{equation}
where we have defined $F^\star(\vl w) \equiv \sum_{\vl n\in\mb Z^d} F\left(\vl w + \frac{2\pi\vl n}\delta\right)$ for a generic function $F$.

The mean-squared error can then be written using the Parseval-Plancherel identity. After some calculations we find:
\begin{multline}
    \mb E\,\mr{MSE} = L^{-d} \mb E\int_{\mc V_d} \mr{d}^d\vl x\, [Z_T(\vl x) - \hat Z_S(\vl x)]^2 = L^{-d} \mb E \sum_{\vl w\in \mb L_d} \left|\tl Z_T(\vl w) - \tl Z_T^\star(\vl w) \frac{\tl K_S(\vl w)}{\tl K_S^\star(\vl w)}\right|^2 = \\
    = L^{-\nicefrac{d}2} \sum_{\vl w \in\mb L_d\cap\mc B_d} \tl K_T^\star(\vl w) - 2 \frac{[\tl K_T\tl K_S]^\star(\vl w)}{\tl K_S^\star(\vl w)} + \frac{\tl K_T^\star(\vl w) [\tl K_S^2]^\star(\vl w)}{\tl K_S^\star(\vl w)^2},\label{eq:msefinal}
\end{multline}
where $\mb L_d = \frac{2\pi}{L}\mb Z^d$ and $\mc B_d = \left[-\frac\pi\delta, \frac\pi\delta\right]^d$ is the Brillouin zone.

In order to simplify this expression in the case where $d_\parallel\leq d$, let us also introduce
\begin{equation}
    F^{{\star_\parallel}}(\vl w_\parallel) \equiv \sum_{\vl n_\parallel\in\mb Z^{d_\parallel}} F\left(\vl w_\parallel + \frac{2\pi\vl n_\parallel}\delta\right).
\end{equation}

Using \eref{eq:ftransparall} it follows that
\begin{equation}
    \tl K_T^\star(\vl w) \propto \delta_{\vl w_\perp} \tl K_T^{\star_\parallel}(\vl w_\parallel),
\end{equation}
\begin{equation}
    [\tl K_T \tl K_S]^\star(\vl w) \propto \delta_{\vl w_\perp} [\tl K_T \tl K_S]^{\star_\parallel}(\vl w_\parallel).
\end{equation}
Plugging the last two equations in \eref{eq:msefinal} we see that, because of the terms $\delta_{\vl w_\perp}$, we find
\begin{equation}
    \mathbb E\,\mr{MSE} \propto \sum_{\vl w_\parallel \in\mb L_\parallel\cap\mc B_\parallel} \tl K_T^{\star_\parallel}(\vl w_\parallel) \left\{1 + \frac{[\tl K_S^2]^\star(\vl w_\parallel)}{\tl K_S^\star(\vl w)^2}\right\} - 2 \frac{[\tl K_T\tl K_S]^{\star_\parallel}(\vl w_\parallel)}{\tl K_S^\star(\vl w_\parallel)}.\label{eq:msedparall}
\end{equation}
Notice that $\tl K^\star_S$ and $[\tl K^2_S]^\star$ do not turn into $[\tl K_S]^{\star_\parallel}$ and $[\tl K_S^2]^{\star_\parallel}$: this is because the Student kernel does not has the same invariants as the Teacher, and it depends on all the components. Here $\mb L_\parallel = \frac{2\pi}{L}\mb Z^{d_\parallel}$, $\mc B_\parallel = \left[-\frac\pi\delta, \frac\pi\delta\right]^{d_\parallel}$.

\emph{(iii) Expansion.}

Using the high-frequency behavior of the Fourier transforms of the two kernels we can write:
\begin{equation}
    \tl K_T^{\star_\parallel}(\vl w_\parallel) \sim \tl K_T(\vl w_\parallel) + \delta^{\alpha_T(d_\parallel)} c_T(d_\parallel) \, \psi_{\alpha_T(d_\parallel)}^\parallel(\vl w_\parallel \delta),
\end{equation}
\begin{equation}
    [\tl K_T \tl K_S]^{\star_\parallel}(\vl w_\parallel) \sim \tl K_T(\vl w_\parallel) \tl K_S(\vl w_\parallel) + \delta^{\alpha_T(d_\parallel) + \alpha_S} c_T(d_\parallel) c_S \, \psi_{\alpha_T(d_\parallel)+\alpha_S}^\parallel(\vl w_\parallel \delta),
\end{equation}
\begin{equation}
    \tl K_S^\star(\vl w_\parallel) \sim \tl K_S(\vl w_\parallel) + \delta^{\alpha_S} c_S \, \psi_{\alpha_S}(\vl w_\parallel \delta).
\end{equation}
We have introduced the functions
\begin{equation}
    \psi_\alpha(\vl w_\parallel) = \sum_{\vl n\in\mb Z^d\setminus\{0\}} \abs{\vl w_\parallel + 2\pi\vl n}^{-\alpha},
\end{equation}
\begin{equation}
    \psi_\alpha^\parallel(\vl w_\parallel) = \sum_{\vl n_\parallel\in\mb Z^{d_\parallel}\setminus\{0\}} \abs{\vl w_\parallel + 2\pi\vl n_\parallel}^{-\alpha}.
\end{equation}
The hypothesis $K_T(\vl 0) \propto \int\mr{d}\vl w\, \tl K_T(\vl w)<\infty$ and $K_S(\vl 0) < \infty$ imply $\alpha_T(d_\parallel) > d_\parallel$ and therefore $\sum_{\vl n_\parallel\in \mb Z^{d_\parallel}} \abs{\vl n_\parallel}^{-\alpha_T(d_\parallel)} < \infty$. We can argue similarly that $\psi_{\alpha_T(d_\parallel)}^\parallel(\vl 0), \psi_{\alpha_T(d_\parallel) + \alpha_S}^\parallel(0), \psi_{\alpha_S}(0)$ are finite. Furthermore, the $\vl w_\parallel$'s in the sums are at most of order $\delta^{-1}$, therefore the terms $\psi_\alpha(\vl w\delta)$ are $\delta^0$ and do not influence how \eref{eq:msefinal} scales with $\delta$.

Expanding \eref{eq:msefinal} and keeping only the highest orders we find:
\begin{multline}
    \mathbb E\,\mr{MSE} \sim \\
    \sim \sum_{\vl w_\parallel\in\mb L_\parallel\cap\mc B_\parallel} \left[2c_T(d_\parallel)\psi_{\alpha_T(d_\parallel)}^\parallel(\vl w_\parallel\delta) \delta^{\alpha_T(d_\parallel)} + c_S^2\psi_{2\alpha_S}(\vl w_\parallel\delta) \frac{\tl K_T^\parallel(\vl w_\parallel)}{\tl K_S^2(\vl w_\parallel)} \delta^{2\alpha_S}\right] + o\left(\abs{\vl w}^{-\alpha_T(d_\parallel) - d_\parallel}\right).\label{eq:scalingsum}
\end{multline}

We have neglected terms proportional to, for instance, $\delta^{\alpha_T(d_\parallel)+\alpha_S}$, since they are subleading with respect to $\delta^{\alpha_T(d_\parallel)}$, but we must keep both $\delta^{\alpha_T(d_\parallel)}$ and $\delta^{\alpha_S}$ since we do not know a priori which one is dominant. The additional term $\delta^{-d}$ in the subleading terms comes from the fact that $|\mb L \cap \mc B| \sim \delta^{-d}$.

The first term in \eref{eq:scalingsum} is the simplest to deal with: since $\abs{\vl w_\parallel\delta}$ is smaller than some constant for all $\vl w_\parallel \in \mb L_\parallel\cap\mc B_\parallel$ and the function $\psi^\parallel_{\alpha_T(d_\parallel)}(\vl w_\parallel\delta)$ has a finite limit, we have
\begin{equation}
\delta^{\alpha_T(d_\parallel)} \sum_{\vl w_\parallel\in\mb L_\parallel\cap\mc B_\parallel} 2c_T(d_\parallel) \psi^\parallel_{\alpha_T(d_\parallel)}(\vl w_\parallel\delta) \sim \delta^{\alpha_T(d_\parallel)} |\mb L_\parallel\cap \mc B_\parallel| \sim \delta^{\alpha_T(d_\parallel)-d_\parallel}.\label{eq:scalingfirstterm}
\end{equation}

We then split the second term in \eref{eq:scalingsum} in two contributions:

\paragraph{Small $\abs{\vl w_\parallel}$} We consider ``small'' all the terms $\vl w_\parallel\in\mb L_\parallel\cap\mc B_\parallel$ such that $\abs{\vl w_\parallel} < \Gamma$, where $\Gamma \gg1$ is of order $\delta^0$ but large. As $\delta\to0$, $\psi_{2\alpha_S}(\vl w_\parallel\delta) \to \psi_{2\alpha_S}(0)$ which is finite because $K_S(0)<\infty$. Therefore
\begin{equation}
\delta^{2\alpha_S} \sum_{\substack{\vl w_\parallel \in\mb L_\parallel\cap\mc B_\parallel\\\abs{\vl w_\parallel}<\Gamma}} c_S^2\psi_{2\alpha_S}(\vl w_\parallel\delta) \frac{\tl K_T^\parallel(\vl w_\parallel)}{\tl K_S^2(\vl w_\parallel)} \to \delta^{2\alpha_S} c_S^2\psi_{2\alpha_S}(0) \sum_{\substack{\vl w_\parallel \in\mb L_\parallel\cap\mc B_\parallel\\\abs{\vl w_\parallel}<\Gamma}} \frac{\tl K_T^\parallel(\vl w_\parallel)}{\tl K_S^2(\vl w_\parallel)}.
\end{equation}
The summand is real and strictly positive because the positive definiteness of the kernels implies that their Fourier transforms are strictly positive. Moreover, as $\delta\to0$, $\mb L_\parallel \cap \mc B_\parallel \cap \left\{\abs{\vl w_\parallel} < \Gamma\right\} \to \mb L_\parallel \cap \left\{\abs{\vl w_\parallel} < \Gamma\right\}$, which contains a finite number of elements, independent of $\delta$. Therefore
\begin{equation}
\delta^{2\alpha_S} \sum_{\substack{\vl w_\parallel \in\mb L_\parallel\cap\mc B_\parallel\\\abs{\vl w_\parallel}<\Gamma}} c_S^2\psi_{2\alpha_S}(\vl w_\parallel\delta) \frac{\tl K_T^\parallel(\vl w_\parallel)}{\tl K_S^2(\vl w_\parallel)} \sim  \delta^{2\alpha_S}.\label{eq:scalingsecondterm1}
\end{equation}

\paragraph{Large $\abs{\vl w}$} ``Large'' $\vl w$ are those with $\abs{\vl w} > \Gamma$: we recall that $\Gamma\gg1$ is of order $\delta^0$ but large. This allows us to approximate $\tl K_T^\parallel$, $\tl K_S$ in the sum with their asymptotic behavior:
\begin{multline}
\delta^{2\alpha_S} \sum_{\substack{\vl w_\parallel \in\mb L_\parallel\cap\mc B_\parallel\\\abs{\vl w_\parallel}>\Gamma}} c_S^2\psi_{2\alpha_S}(\vl w_\parallel\delta) \frac{\tl K_T^\parallel(\vl w_\parallel)}{\tl K_S^2(\vl w_\parallel)} \propto \delta^{2\alpha_S} \sum_{\substack{\vl w_\parallel \in\mb L_\parallel\cap\mc B_\parallel\\\abs{\vl w_\parallel}>\Gamma}} \abs{\vl w_\parallel}^{-\alpha_T(d_\parallel)+2\alpha_S} \approx \\
\approx \delta^{2\alpha_S} \int_{\Gamma}^{\nicefrac1\delta} \mr d w_\parallel\, w_\parallel^{d_\parallel-1-\alpha_T(d_\parallel)+2\alpha_S} \sim \delta^{\min(\alpha_T(d_\parallel)-d_\parallel, 2\alpha_S)}.\label{eq:scalingsecondterm2}
\end{multline}

Therefore in the end
\begin{equation}
    \mb E\,\mr{MSE} \sim \delta^{\min(\alpha_T(d_\parallel) - d_\parallel, 2\alpha_S)} \equiv \delta^{\beta d}.
\end{equation}

The kernels $K$ that we consider in the present article, namely Laplace and Mat\'ern, share the property that the respective exponents take the form $\alpha_K(d) = d + \theta_K$, $\theta_K$ being a dimension-independent constant that only depends on the isotropic function that defines the kernel. For instance, we have $\alpha=d+1$ for Laplace and $\alpha(d) = d + 2\nu$ for Mat\'ern (with parameter $\nu$). Consequently, for these kernels the term $\alpha(d_\parallel) - d_\parallel$ that appears in the last equation is actually independent of $d_\parallel$, and therefore so is the exponent $\beta$. We believe that this structure of the exponent $\alpha(d)$ is more general. Signals that point in this direction can be found in several papers. In (\mycite{grafakos2013fourier}) they show that (with our notation), for functions $K\left(\abs{\vl x}\right)$ that are integrable in $\mb R^d$ and $\mb R^{d+2}$,
\begin{equation}
    \mc F_{d+2}\left[K\left(\abs{\vl x}\right)\right](w) \propto w^{-1}\partial_w \mc F_d\left[K\left(\abs{\vl x}\right)\right](w),
\end{equation}
and so if the Fourier transform in dimension $d$ decays as $w^{-\alpha(d)}$, in dimension $d+2$ it decays with an exponent $\alpha(d+2) = \alpha(d) + 2$. In (\mycite{estrada2014radial}) they prove a result for functions belonging to the Schwartz space (rapidly decreasing functions). This result implies that if the Fourier transform in dimension $d+1$ decays with an exponent $\alpha(d+1)$, then in dimension $d$ the function decays with the exponent $\alpha(d) = \alpha(d+1) - 1$. 

These results offer a link between the exponents in different dimensions. In (\mycite{erdelyi1955asymptotic}) the author computes the asymptotic behavior of the one-dimensional Fourier transform of functions with a singularity. In particular, it follows that if $K(x) = |x|^{\theta_K} K_\infty(x)$, with $-1<\theta_K\leq0$ and $K_\infty\in C^\infty(\mb R)$, then its Fourier transform at the leading order decays with an exponent $\alpha(d=1)=1+\theta_K$. There is a similarity with the value of the exponents for the Laplace and Mat\'ern kernels that we use: the value of $\theta_K$ is linked to the exponent of the cusp $|x|^{\theta_K}$ that appears in the Taylor expansion of the Kernel at the origin. We expect that this fact, namely that the exponent $\alpha_K(d)$ is the sum of spatial dimension $d$ and of the cusp exponent $\theta_K$, is more generic and applies to most of the kernels that are used in practice.

\section{Regime \texorpdfstring{$\sigma\ll\delta$}{large sigma}: curse of dimensionality}\label{app:smallsigma}

We consider here the case where the kernel bandwidth $\sigma$ is much smaller than the nearest-neighbor distance $\delta$. In this limit the contributions in the expansion of the decision boundary in \eref{eq:svdecbound} are significantly suppressed because the kernel is supposed to decay when its argument is large, and the decision boundary is dominated by the charge of training pattern $\vl x^\mu$ that is closest to $\vl x$. The sign of the decision function is thus fixed by the sign of the nearest neighbor's charge and the accuracy is driven by the nearest neighbor distance, which is susceptible to the curse of dimensionality.

We can see this more precisely if we approximate the kernel interaction between two points $\vl x$ and $\vl x^\prime$ as
\begin{equation}
K\left(\frac{\abs{\vl x-\vl x^\prime}}{\sigma}\right) \approx
\begin{cases}
a_0 = K(0) \text{ if  } \vl x = \vl x^\prime,\\
a_1 = K\left(\frac\delta\sigma\right) \ll a_0 \text{ if } \vl x^\prime \text{ is one of the nearest neighbors of } \vl x \footnotemark,\\
0 \text{ otherwise}.
\end{cases}
\end{equation}
\footnotetext{For the derivation of the following scalings the notion of ``nearest neighbors'' could be relaxed to include points that lie in a thin shell. In any case we assume that the number of nearest neighbors of a given point if finite.}
Hence, the decision function at a point $\vl x^\mu$ reads
\begin{equation}
    f(\vl x^\mu) \approx a_0 \alpha^\mu y^\mu + a_1 \sum_{\nu \in \partial \vl x^\mu} \alpha^\nu y^\nu + b \approx (a_0 + a_1^\prime) (\alpha_0 + y^\mu \Delta\alpha) y^\mu + b,\label{eq:approxboundary}
\end{equation}
where the sum runs over the nearest neighbors of $\vl x^\mu$. We use that all points are SV, which results from the hierarchy $a_1 \ll a_0$. Indeed, the interaction term alone is never sufficient for $\abs{f(\vl x^\mu)}$ to exceed one. The second equality is justified by the following reasoning. First, in the limit $\delta\to0$, the nearest neighbors typically share the same sign, so that all the $y^\nu$'s in the sum can be replaced by $y^\mu$. $a_1^\prime$ is thus $a_1$ times the number of terms in the sum. Then, because the distribution is assumed smooth and the kernel is blind to the data structure coming from distant patterns, the SV charge may only depend on its label: $\alpha^\mu = \alpha_0 + y^\mu \Delta\alpha$. $\Delta\alpha$ is taken independent of the associated label $y^\mu$, as we assume the labels to be balanced. The charge conservation \eref{eq:chcons} implies immediately that $\Delta\alpha = -\alpha_0\<y\>$, where $\<y\> = \frac1p \sum_\mu y^\mu \sim p^{-\nicefrac{1}{2}}$ and imposing the condition $y^\mu f(\vl x^\mu) = 1$ on each points $\vl x^\mu$ yields $\alpha_0 = 1/(a_0 + a_1')$ and $b=\<y\>$.

\begin{figure}[ht]
\centering
\begin{subfigure}{.49\textwidth}
  \centering
  \includegraphics[width=1\linewidth]{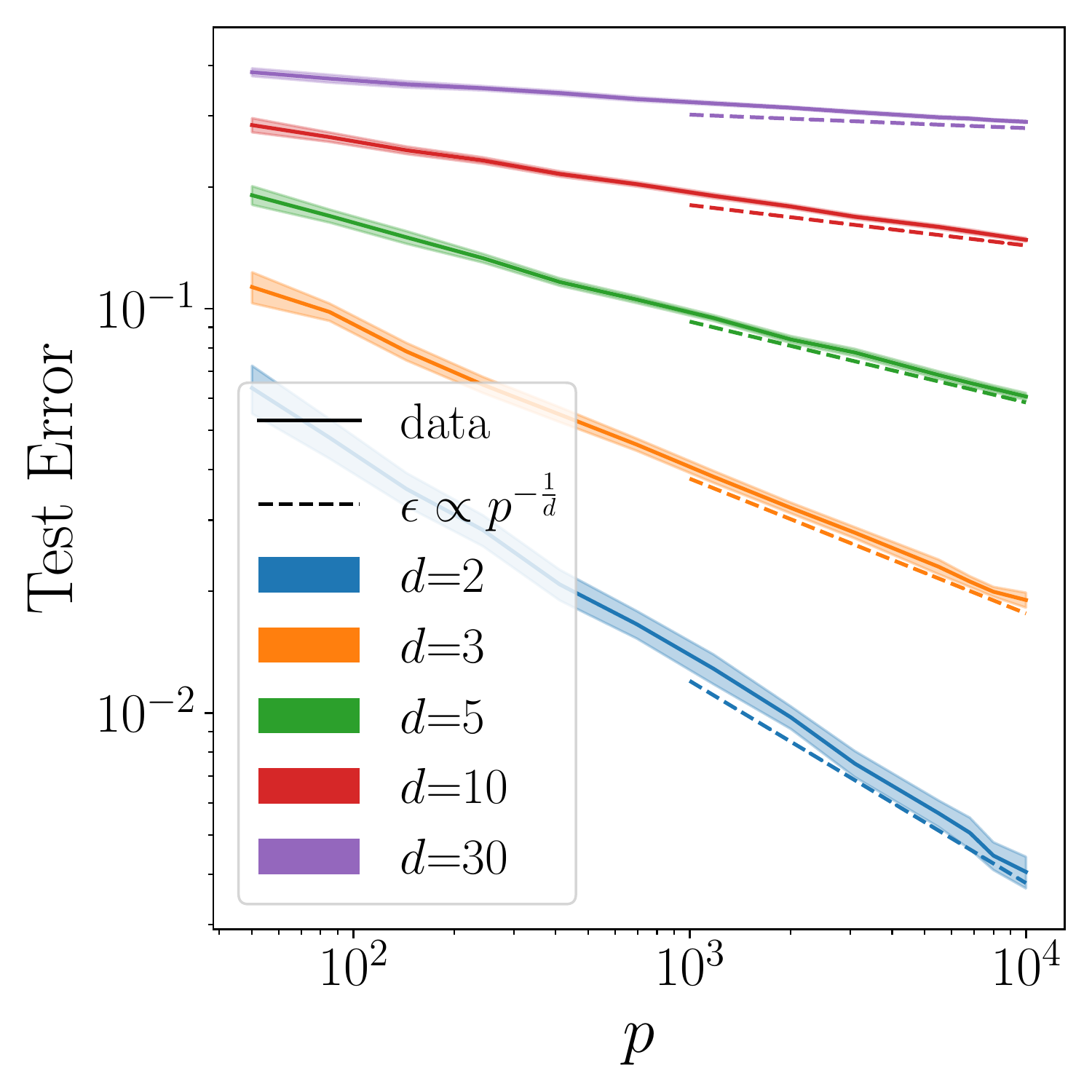}
\end{subfigure}
\begin{subfigure}{.49\textwidth}
  \centering
  \includegraphics[width=1\linewidth]{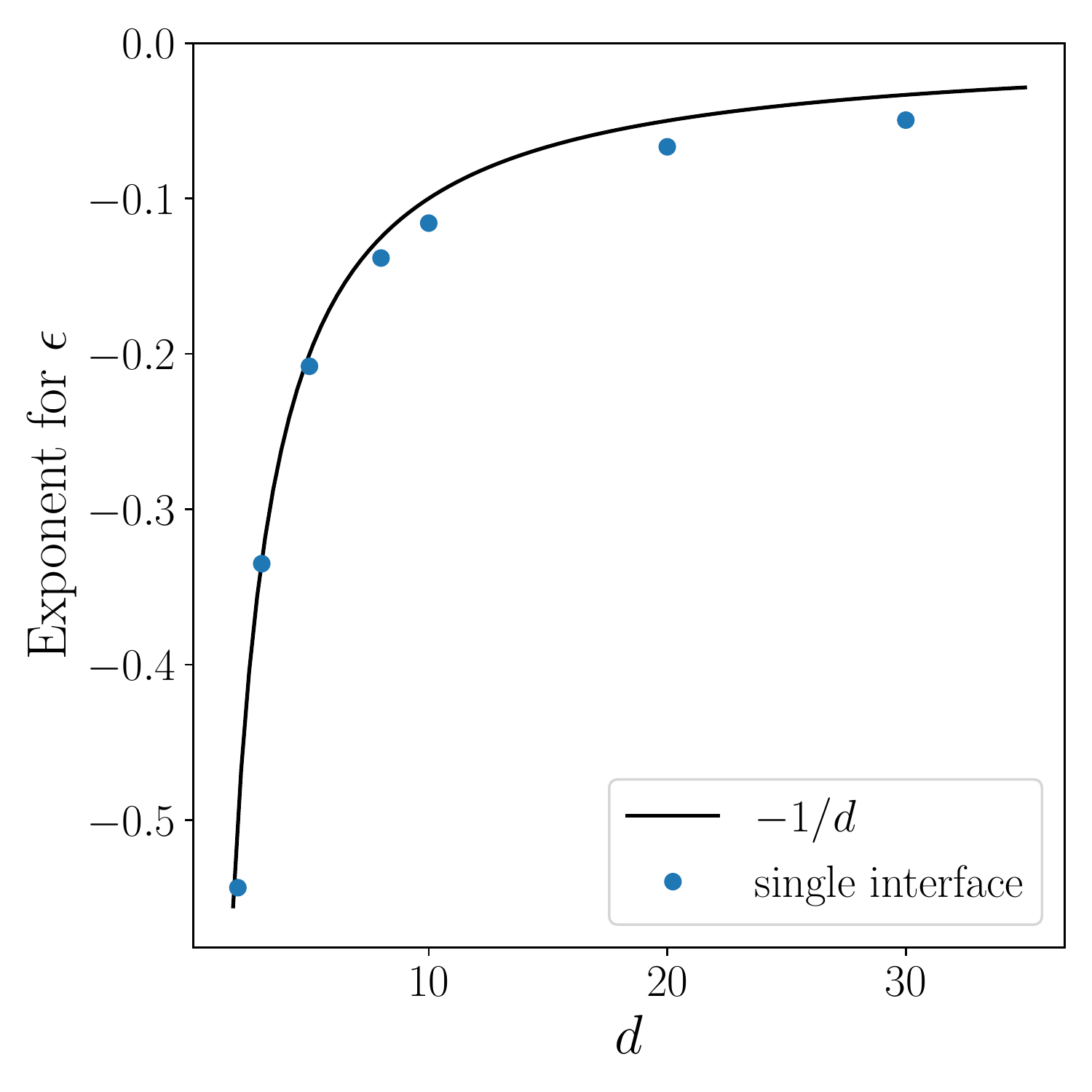}
\end{subfigure}
\caption{\label{fig:small_sigma} \underline{Left}: Test error vs the size of the training-set size $p$ for the \textbf{single-interface setup} in the vanishing bandwidth regime. The points in the dataset are drawn from the standard normal distribution in dimension $d$ (see the color legend) and learned with the margin-SVC algorithm with the Laplace kernel ($\xi=1$) of bandwidth $\sigma=10^{-2}$. The solid lines correspond to the average over 50 initializations, while the shaded regions are the associated standard deviations. The dashed lines illustrate the power law $\epsilon\sim p^{-1/d}$. The bias of the SVC decision function has been removed by hand to avoid that the test error remains stuck at 50\% as discussed at the end of \aref{app:smallsigma}. \underline{Right}: The power law exponents are extracted by fitting the curves on the left plot and compared to the nearest neighbor prediction.}
\end{figure}

We can now compute the test error of the SVC in the limit $\sigma\ll\delta$. The prediction on a test point $\vl x$ is
\begin{equation}
    \hat y(\vl x) \approx \mr{sign}\left( a_1 \sum_{\nu \in \partial \vl x} \alpha^\nu y^\nu + b\right) \approx \mr{sign}\left[ \frac{a_1^\prime}{a_0}y_\mr{NN} + b \right],
\end{equation}
where with a slight abuse of notation we take the sum over the points $\vl x^\nu$ in the training set that are nearest neighbors of the test point $\vl x$, and $y_\mr{NN}$ is their label (as before, assumed to be constant among nearest neighbors). We observe two distinct behaviors according to the ratio between the bias $b=\<y\>$ and the nearest-neighbor contribution $a_1^\prime$. If $\<y\> \sim p^{-\nicefrac{1}{2}}$ is much larger than $a_1^\prime$, the above prediction yields $\hat y(\vl x) = \mr{sign}\,\<y\>$ (for any $\vl x$): this estimator cannot beat a 50\% accuracy. On the contrary, if $\<y\>$ is much smaller than $a_1^\prime$, the prediction yields $\hat y(\vl x) = \mr{sign}(y_\mr{NN})$: the classifier acts as a nearest-neighbor algorithm, and consequently its test error scales as the nearest-neighbor distance, $\epsilon \sim \delta \sim p^{-1/d}$ --- namely, it is susceptible to the curse of dimensionality --- as we show in figure \fref{fig:small_sigma}.

\section{Proof that power kernels are CSPD}\label{app:proof_cspd}

The margin-SVC algorithm presented on \sref{sec:large_sigma} relies on the assumption that the Gram matrix is conditionally strictly positive definite (CSPD). In this appendix, we prove that the power kernel $K(\vl x, \vl x^\prime) = -\big(\frac{\abs{\vl x - \vl x^\prime}}{\sigma}\big)^\xi$ indeed belongs to the CSPD class for $0<\xi<2$ and for any space dimension, by introducing the following definitions and theorems:

\paragraph{Definition:} \emph{A real function $k$ is called conditionally strictly positive definite (CSPD) in $\mb R^d$, if}
\begin{equation}
    \sum_{\mu=1}^p \sum_{\nu=1}^p c_\mu c_\nu k\left(\abs{\vl x_\mu-\vl x_\nu}\right) > 0,
\end{equation}
\emph{for any set of $p$ distinct points $\vl x_1, \dots, \vl x_p \in \mathbb{R}^d$ and any choice of $p$ variables $c_1, \dots, c_p$, satisfying}
\begin{equation}
    \sum_{\mu=1}^p c^\mu = 0.
\end{equation}

\paragraph{Definition:} \emph{A function $\phi$ is said completely monotone in $(0, \infty)$ if is satisfies $\phi \in C^\infty(0, \infty)$ and $(-1)^n \partial^{(n)} \phi(r) \geq 0$, for all $n \in \mathbb{N}_0$ and all $r>0$.}

\paragraph{Theorem:} \emph{Let $\phi \in C[0, \infty) \cap C^\infty(0, \infty)$. The function $k(\bullet) = \phi(\abs{\bullet}^2)$ is CSPD in $\mb {R}^d$ for all $d$, if and only if its negative derivative $-\phi^\prime$ is completely monotone on $(0, \infty)$ and $\phi$ is not a polynomial of degree at most one. A proof can be found in chapter 8 of (\mycite{wendland2004scattered}).}\vspace{0.5em}

The introductory statement arises naturally when considering the univariate function $\phi(r) = -r^{\xi/2}$ defined on $\mathbb{R}_+$. Following the theorem and the definitions, one easily show that the function $-\phi^\prime(r) = \frac{\xi}{2}r^{\xi/2-1}$ is completely monotone on $(0, \infty)$ for $0\leq\xi\leq2$. The condition that $\phi$ be not a polynomial of degree at most one excludes further the cases $\xi=0$ and $\xi=2$, which proves that the function $k(r) = -r^\xi$ is CSPD for $0<\xi<2$. Note that a radial kernel is defined as the multivariate function $K(\vl x, \vl x^\prime) = k\left(\abs{\vl x - \vl x^\prime}\right)$, and that if the kernel generator $k$ is CSPD, the kernel $K$ is also called CSPD.

\section{Large \texorpdfstring{$\sigma$}{sigma} convergence of the SVC algorithm}\label{app:infsigmalimit}

In section \ref{sec:large_sigma}, it is loosely argued that in the limit of large $\sigma$ one could replace the actual kernel $K(r/\sigma)$ by its truncated Taylor expansion $\hat K(r/\sigma)$. Here, we prove that in the limit $\sigma \to \infty$, the SVC solution with the truncated kernel converges to the actual SVC solution: $\{\hat \alpha^\mu\} \xrightarrow{\sigma\to\infty} \{\alpha^\mu\}$. 

We assume that the kernel $K$ can be written as:
\begin{equation*}
    K\left(\frac{r}{\sigma}\right) = \hat K\left(\frac{r}{\sigma}\right) + o\left(\sigma^{-\xi}\right) \,\, \mr{,with} \,\, \hat K\left(\frac{r}{\sigma}\right) = c_0 + c_1 \left(\frac{r}{\sigma}\right)^\xi 
\end{equation*}
For a given classification problem $\{(\vl x^\mu, y^\mu)\}$, the SVC algorithm converges to a set of dual variables $\{\alpha^\mu\}$, respectively $\{\hat \alpha^\mu\}$ provided that the associated kernel is conditionally strictly positive definite (CSPD). $\hat K$ is proved to be CSPD in appendix \aref{app:proof_cspd} if $c_1 < 0$ and $0<\xi<2$, while $K$ is assumed to be CSPD from the start. This condition guarantees that the Lagrangian in \eref{eq:SVC_max_equation} defines a strictly convex problem. Rescaling the dual variables $\alpha^\mu \to \alpha^\mu / \sigma^\xi$ yields the following rescaled Lagrangians:
\begin{equation}
    \hat{\mc L}(\alpha) = \sum_{\mu=1}^p \alpha^\mu - \frac{c_1}{2} \sum_{\mu,\nu=1}^p \alpha^\mu \alpha^\nu y^\mu y^\nu \abs{\vl x^\mu - \vl x^\nu}^\xi \,\, \mr{and} \,\, 
    \mc L(\alpha) = \hat{\mc L}(\alpha) + \epsilon(\sigma),
\end{equation}
The rescaled solution $\{\hat \alpha^\mu\}$ of the maximizing problem with the Lagrangian $\hat{\mc L}$ is well defined in the limit $\sigma\to\infty$, hence the strict convexity of both Lagrangian ensures that $\{\hat \alpha^\mu\} \to \{\alpha^\mu\}$, when the perturbation $\epsilon(\sigma)$ vanishes.

\section{The charge structure factor}\label{app:charge_structure}

\begin{figure}[ht]
\centering
\begin{subfigure}{.49\textwidth}
  \centering
  \includegraphics[width=1\linewidth]{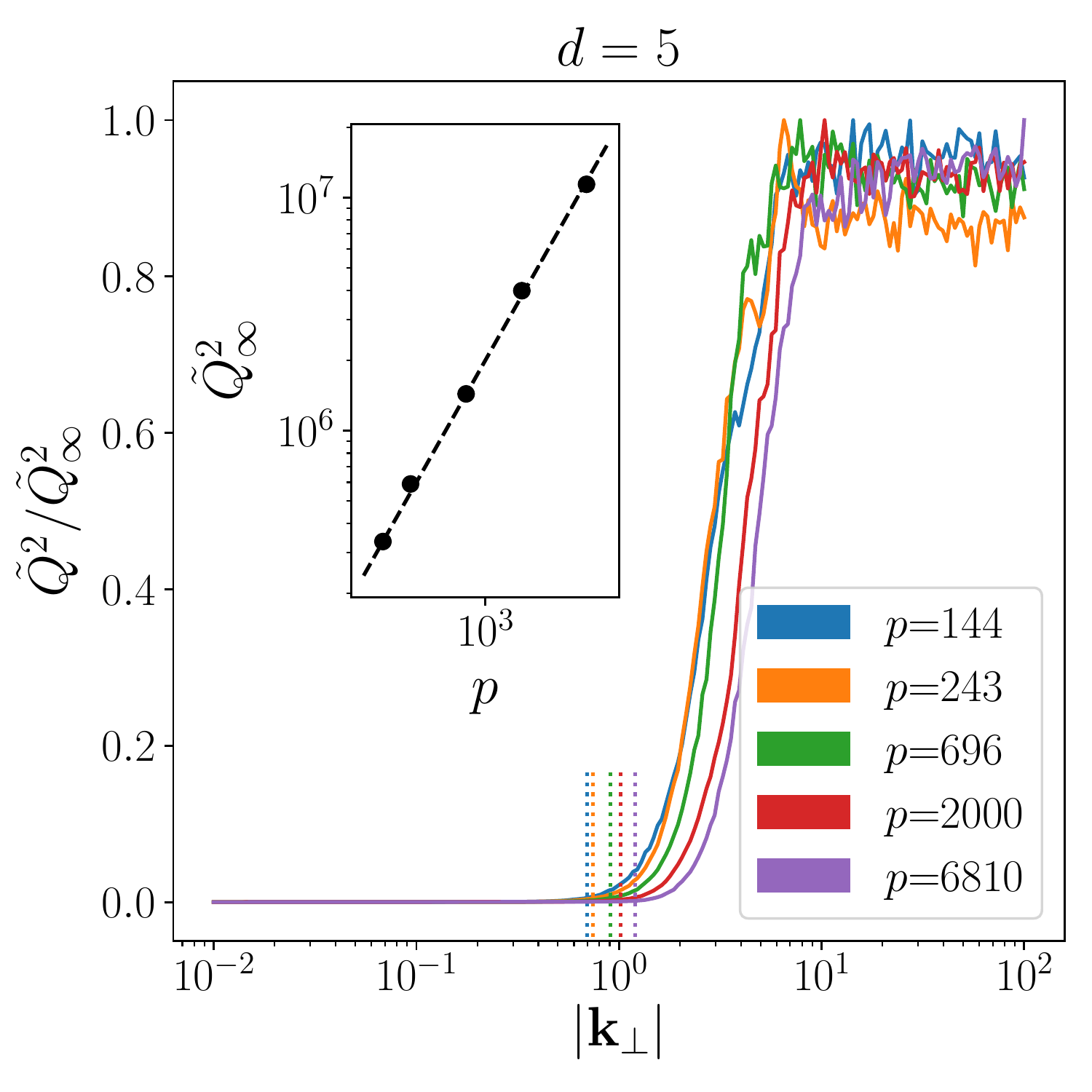}
\end{subfigure}
\begin{subfigure}{.49\textwidth}
  \centering
  \includegraphics[width=1\linewidth]{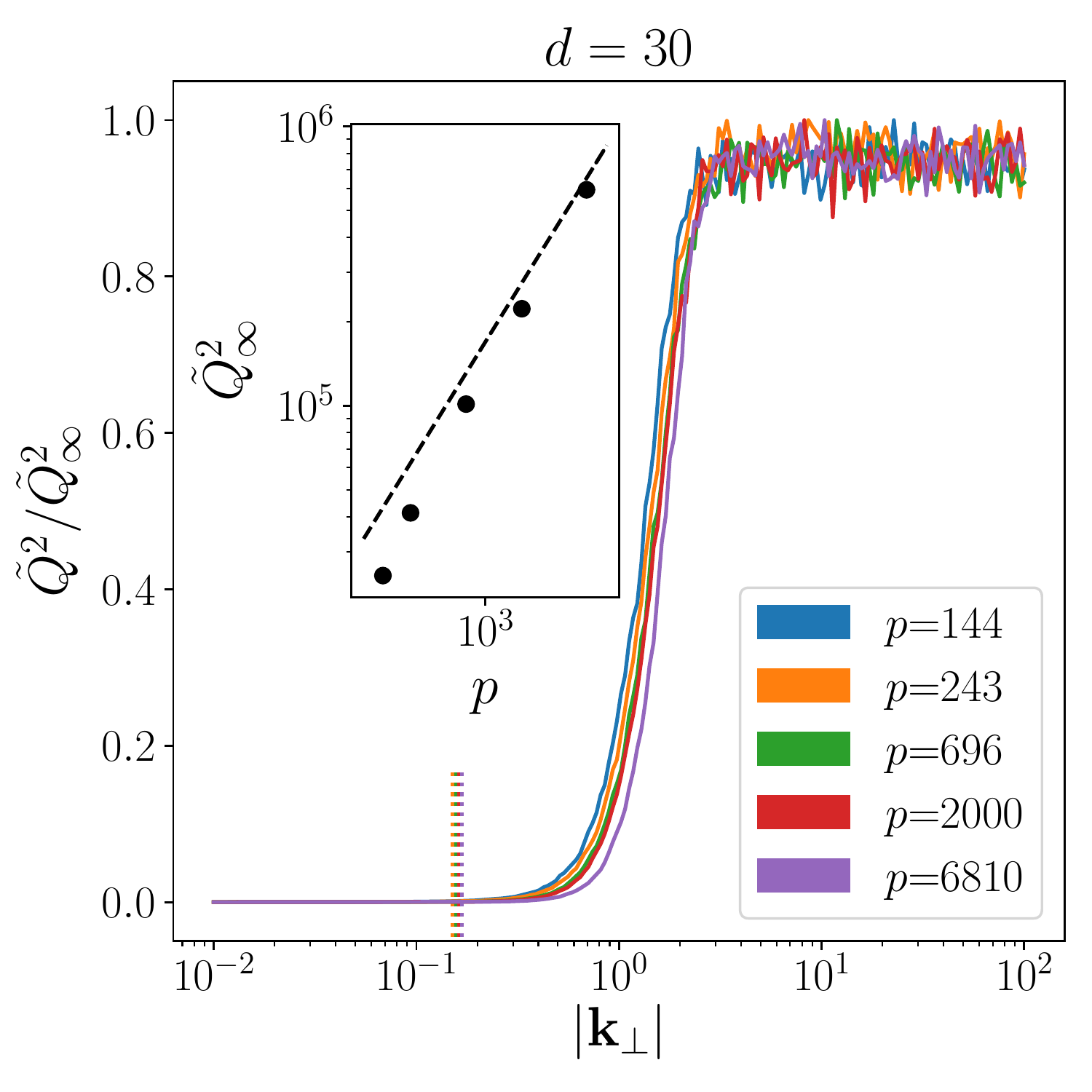}
\end{subfigure}
\caption{\label{fig:charge_structure_factor} Charge structure factor as a function of the (transverse) wave vector amplitude $\abs{\vl k_\perp}$, for different training set sizes $p$ and dimensions $d=5,30$. We plot the square $\tl Q^2\left(\vl k_\perp\right)$ averaged over $N=2000$ samples, normalized by the expected high-frequency variance $\tl Q^2_\infty = \bar\alpha^2 p \Delta/\gamma$. The inset plot shows $\tl Q_\infty
^2$ vs the size of the training set $p$.}
\end{figure}

The charge structure factor $\tl Q$ introduced in \eref{eq:f_fourier_transform} is a good measure of the fluctuations in the system and, in particular, of the cutoff occurring at the scale $r_c$. It is argued in \sref{sec:large_sigma} that $\tl Q^2(\vl k_\perp) \sim \bar\alpha^2 p \Delta / \gamma$ at large frequencies, namely $\abs{\vl k_\perp} > r_c^{-1}$. This scaling is verified numerically in \fref{fig:charge_structure_factor}.

The data are obtained as follows: for each $\abs{\vl k_\perp}$, a set of $N=2000$ random wave vectors are generated on the interface; the associated factor is computed by summing over the SV of the considered setup and then averaged. The fluctuations observed at large $\abs{\vl k_\perp}$  decrease when $N$ increases. The insets illustrate the expected asymptotic behavior $\tl Q^2_{\infty} \approx \bar\alpha^2 p \Delta/\gamma$, while the vertical dotted lines correspond to the typical nearest-neighbor distance $r_c$.

\section{SVC gradient of the multiple-interfaces setup}\label{app:multiple_interfaces}

Consider a setup of $n$ (odd) interfaces separated by a distance $w$. We count the interfaces with the index $I=-\frac{n-1}{2}, \dots, \frac{n-1}{2}$ and set the middle interface at $x_1=0$, so that $x_{1,I} = I w$. We call $\Delta_I$ the band thickness on each side of the $I$st interface and denote the mean SV dual variable on its left, respectively on its right, by $\bar\alpha_I$, respectively $\bar\alpha^\prime_I$. Without loss of generality, we fix the sign of the setup by setting $y(x_1) = +1$, for $x_1\in [0, w]$. The symmetry of the system imposes that $\Delta_{-I} = \Delta_I$ and $\bar\alpha^\prime_I = \bar\alpha_{-I}$ for all $I>0$, as well as that $\bar\alpha^\prime_0 = \bar\alpha_0$.

Following the same construction as in \sref{sec:large_sigma}, in the central-limit approximation the SVC function on the point $\vl x = (x_1, \vl 0)$ is given by
\begin{equation}\label{eq:mult_interface_SVC}
    f(x_1) = b - p \sigma^{-\xi} \gamma^{-d} \, \sum_{I=-\frac{n-1}{2}}^{\frac{n-1}{2}} (-1)^I \, \int_{-\Delta_I}^{\Delta_I} \mr{d}u \, \mr{sgn}(u) \, \bar\alpha_I(u) \, g(x_1-x_{1,I}-u),
\end{equation}
where
\begin{equation}
    g(x) = \int \mr{d}\vl x_\perp (x^2 + \abs{\vl x_\perp}^2)^{\xi/2} \sim \underbrace{\int_0^w \mr{d}r  \,r^{d-2} \, (x^2 + r^2)^{\xi/2}}_{g_S(x)} + \underbrace{\int_w^\gamma \mr{d}r \, r^{d-2} \, (x^2 + r^2)^{\xi/2}}_{g_L(x)},
\end{equation}
and $\bar\alpha_I(u) = \alpha_I$, respectively $\bar\alpha_I(u) = \alpha^\prime_I$, for $u<0$, respectively $u>0$. By symmetry, the target function is of the form\footnote{The shift constant $\beta_0$ is discarded because of the bias freedom in \eref{eq:mult_interface_SVC}.}
\begin{equation}
    f(x_1) = \beta_1 x_1 + \cdots + \beta_i x_1^i + \cdots + \beta_n x_1^n+O(x_1)^{n+2},
\end{equation}
with $i$ only running over odd indices. Imposing that the target function is zero on each interface, all coefficients can be expressed in terms of $\beta_n$: $\beta_i = b_i w^{n-i} \beta_n$, where $b_i \sim \mc{O}(1)$. Similarly the SVC condition that $\partial_{x_1}f(x_I) \Delta_I$ is identical on each interface, allows to relate all band thicknesses to $\Delta_0$: $\Delta_I = d_I \Delta_0$, with $d_I \sim \mc{O}(1)$. Denote by $\alpha$, $\delta\alpha$ and $\Delta$ respectively the typical value of $(\alpha^\prime_I + \alpha_I)/2$, $\abs{\alpha^\prime_I - \alpha_I}/2$ and $\Delta_I$. One can obtain the $\beta$ coefficients associated to \eref{eq:mult_interface_SVC} by differentiating it, namely
\begin{equation}
    \beta_i = \frac{f^{(i)}(0)}{i!} = \underbrace{\sum_{j=0}^\infty g^L_{i+1+2j} T_j}_{\beta^L_i} + \underbrace{g^S_{i+1} w^{d+\xi-2-i} \Delta^2 \alpha}_{\beta^S_i},
\end{equation}
 where $g^L_i \sim \gamma^{\xi-1-i} \sigma^{-\xi}$, $g^S_i \sim \gamma^{-d} \sigma^{-\xi}$ and $T_j \sim \mc O\left(\Delta^2 w^{2j} \alpha\right) + \mc O\left(\Delta w^{2j+1} \delta\alpha\right)$. The constrained scaling between the $\beta$ coefficients forces the terms of index $j=0,\dots,n-i$ in the sum defining $\beta^L_i$ to cancel each other up to higher order. In particular, $\delta\alpha \sim \alpha \Delta/w$, and $\beta^L_n \sim w^2 \Delta^2 \alpha$. Eventually, the scaling of the gradient depends on the hierarchy between $\beta^L_n$ and $\beta^S_n$:
 \begin{equation}
1 \sim \Delta \partial_{x_1} f \sim 
\begin{cases}
p \, \alpha \,\left(\frac{\gamma}{\sigma}\right)^\xi \, \left(\frac{\Delta}{\gamma}\right)^3 \, \left(\frac{w}{\gamma}\right)^{n+1}, &\mr{ if } \, n \geq d + \xi - 4, \\
p \, \alpha \,  \left(\frac{w}{\sigma}\right)^\xi \, \left(\frac{\Delta}{w}\right)^3 \, \left(\frac{w}{\gamma}\right)^{d}, &\mr{ if } \, n \leq d + \xi - 4.
\end{cases}
 \end{equation}
Also, if $n > d + \xi -1$, when computing $\beta^S_n$, divergences will occur while differentiating $g^S$. This sets an upper bound on the number of interfaces we can consider without considering microscopic effects on the gradient. For an even number of interfaces, a similar discussion holds with the difference that $n$ should be replaced by $n-1$ in the above expressions. Finally, the resulting scaling of the usual observables are given in \sref{sec:multiple_interfaces}.

\section{SVC scaling with the Matérn kernel}\label{app:matern}

\begin{figure}[H]
\includegraphics[scale=0.64]{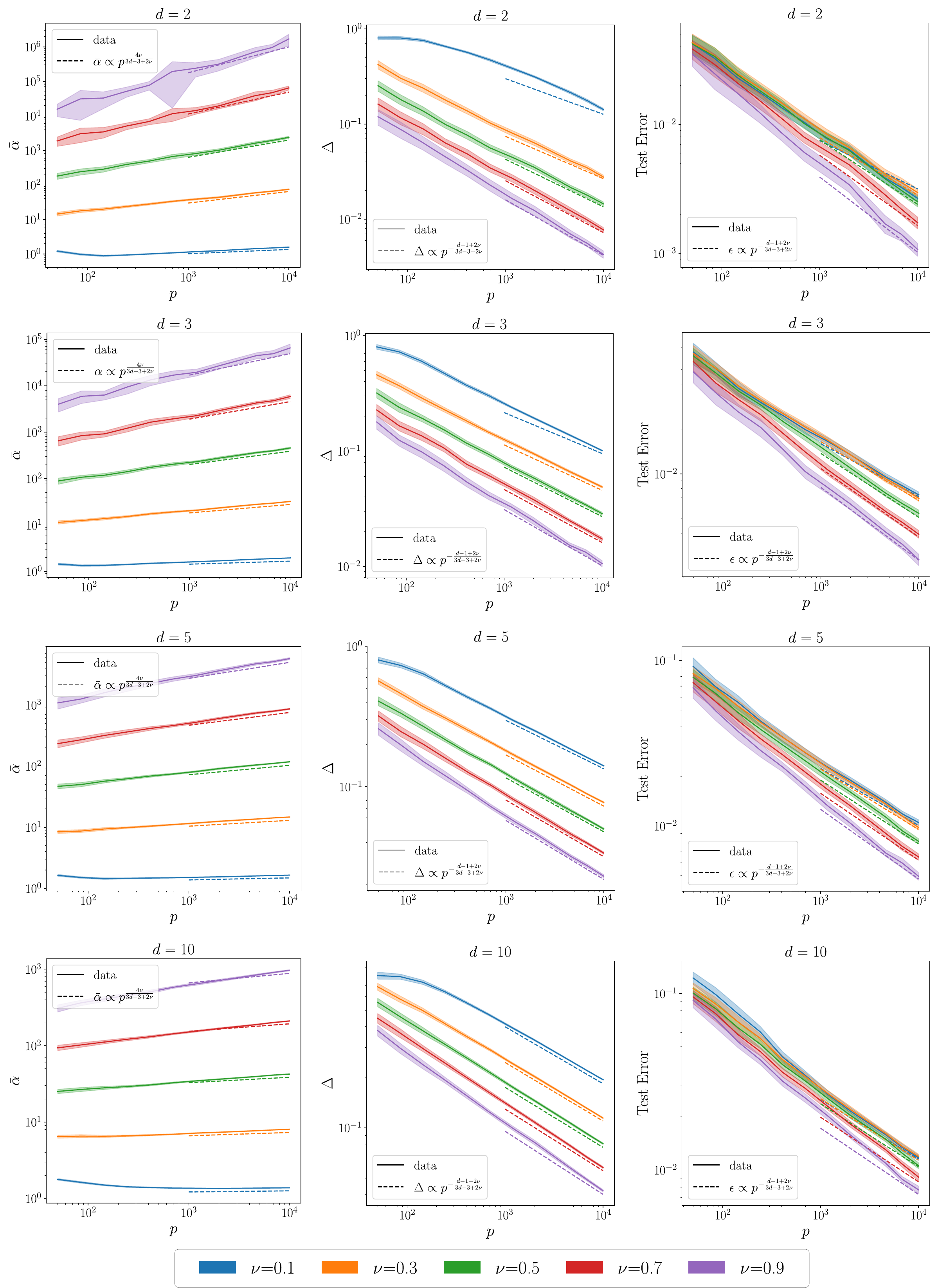} 
\caption{Dependence on the training set size $p$ of the SV mean dual variable $\bar\alpha$ (left), the SV band thickness $\Delta$ (middle) and the test error (right) for the \textbf{single-interface setup} in dimensions $d=2, 3, 5, 10$. The SVC algorithm is run with the Matérn kernel \eref{eq:matern_kernel} with bandwidth $\sigma=100 \gg \delta$ and parameter $\nu=0.1, 0.3, 0.5, 0.7, 0.9$ for which the kernel is conditionally strictly positive definite. The solid lines are averaged over 50 initializations and the shaded regions represent the standard deviation. Dashed lines illustrate the power-law predictions of \eref{eq:power_laws} and \eref{eq:scalelargesigma}.}
\label{fig:matern_scaling}
\end{figure}

Results of simulations on the single-interface setup with Mat\'ern kernels are shown in \fref{fig:matern_scaling}, for several parameters $\nu$ and several dimensions $d$. All the curves follow the scalings predicted in \sref{sec:large_sigma}.

\section{Numerical definition of the scale \texorpdfstring{$r_c$}{rc}}\label{app:critical_scale}
\begin{figure}[H]
\centerline{\includegraphics[scale=.4]{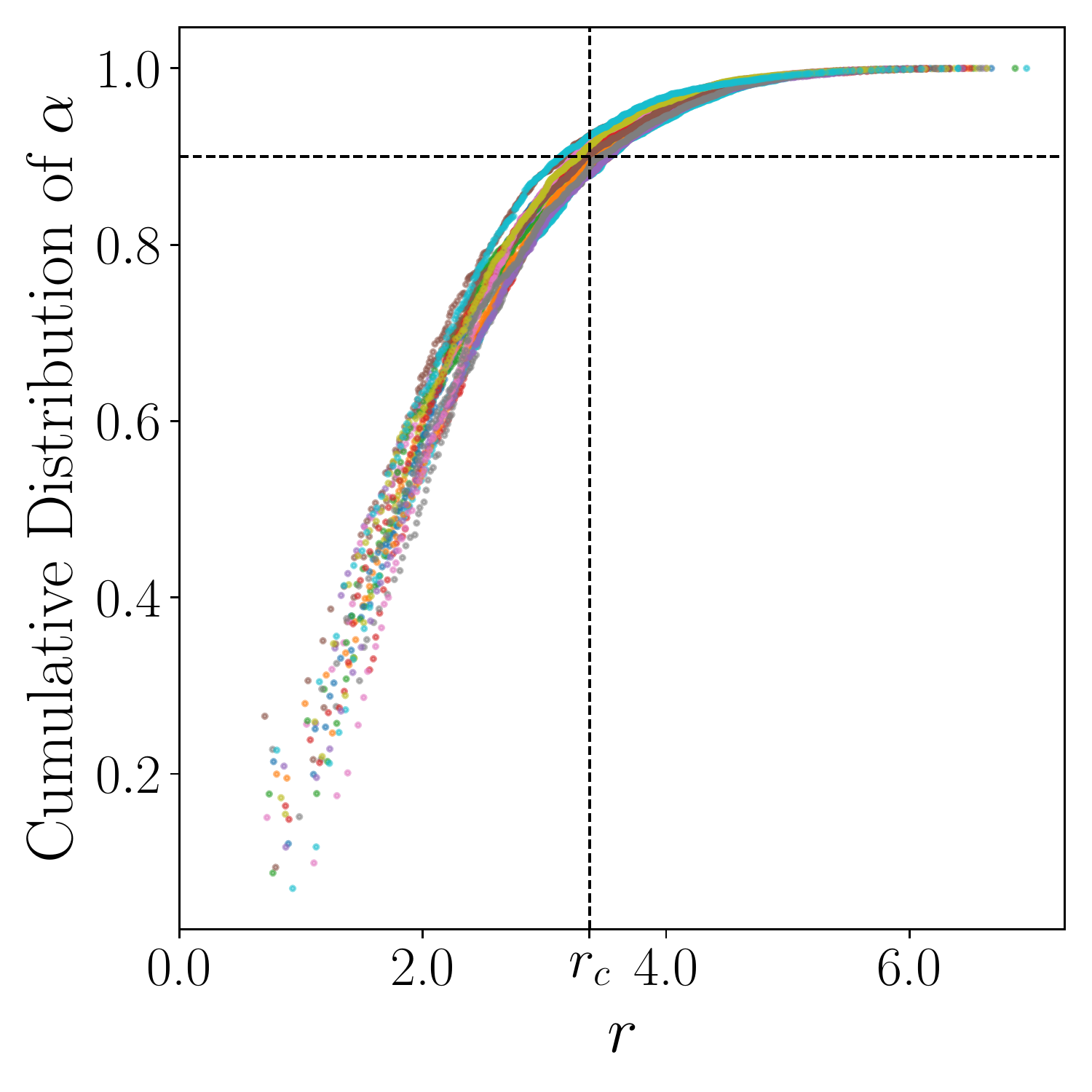}}
\caption{\label{fig:rc_definition} Example of the cumulative distribution of the amplitude of the dual variable variation as a function of the distance $r$ to the additional point (see the text above). Each color corresponds to a different realization of the interface setup with $d=5$ and $p=6810$. The vertical dashed line stands for the scale $r_c$ averaged over all realizations obtained with $C=0.9$ (horizontal dashed line).}
\end{figure}

In \sref{sec:large_sigma}, the scale $r_c$ is defined geometrically as the distance between nearest support vectors. The numerical definition of $r_c$ is different as it aims at confirming the ``minimal disturbance hypothesis'' presented in the note at the end of \sref{sec:large_sigma}. From this point of view, the scale $r_c$ is also the scale behind which the charge of two SVs are not correlated. To test this idea, the solution of the margin-SVC problem is computed once for a benchmark training set and a second time for the same training set with one additional point close enough to the interface to be a SV. We then calculate the cumulative distribution of the charge variations $d\alpha^\mu = \abs{\alpha^\mu-\alpha^{\prime\mu}}$ as function of their distance to the additional point $r^\mu$. The resulting distribution is displayed on \fref{fig:rc_definition} for multiple realizations of the single interface setup with $d=5$ and $p=6810$. The scale $r_c$ is then defined as the distance for which the cumulative distribution reaches a given value $C < 1$. The particular choice of $C$ doesn't alter the power law behavior.

\section{Scaling arguments for the spherical setup}\label{app:spherescalings}

In this appendix, we sketch how the scaling relations in \sref{sec:large_sigma} may be derived for the spherical interface setup discussed in \sref{sec:sphere}, where the label only depends on the norm of the vector: $y(\vl x) = \text{sign}(\abs{\vl x} - R)$, with $R$ the radius of the sphere. In the same line as for the linear interface, it is assumed that all SVs lie within a shell of thickness $\Delta \ll R$ around the interface. The decision function on the vector $\vl x$,
\begin{equation}
    f(\vl x) =  b - \sum_{\mu=1}^p \alpha^\mu y^\mu \left(\frac{\abs{\vl x - \vl x^\mu}}{\sigma}\right)^\xi,
\end{equation}
is better apprehended in a Cartesian frame such that $\vl x = (x_1 = \abs{\vl x}, \vl 0)$, which requires to rotate all SVs: $\vl x^{\mu} \to x'^{\mu} = \mathcal{R} \vl x^{\mu}$. In the large $p$ limit, the charge conservation, $Q=\sum_{\mu=1}^p \alpha^{\mu} y^{\mu}=0$, reads
\begin{equation}
    0 = \int \mr{d}^{d}\vl x \rho(\vl x) \alpha(\vl x) y(\vl x) = S_{d-1} \int_{-\Delta}^{\Delta} \mr{d}u (R+u)^{d-1} \rho(R + u) \alpha(R+u) y(R+u). \label{eq:sphere_charge_conservation}
\end{equation}
Spherical coordinates are used in the second equality: the angular variables trivially integrate to the unit $(d-1)$-sphere surface, $S_{d-1}$, and the variable $u = r - R$ is used instead of the radius $r = \abs{\vl x}$. For simplicity, we assume that the population distribution is radial: $\rho(\vl x) = \rho(r)$. Were it not the case, the angular integral would merely yield a different finite factor. 

As for the linear interface, the first scaling relation stems from the condition $\Delta \cdot \partial_{x_\parallel} f(\vl x^\star) \sim 1$, for any $\vl x^\star$ lying on the spherical interface. According to the change of frame introduced above, the relevant direction correspond to the first coordinate, namely $\vl x_\parallel = \vl x_1$. The gradient expression \eqref{eq:svcgradient} can thus be expressed as an integral in spherical coordinate with the north pole $\vl x^\star = (R, \vl 0)$:
\begin{equation}
    \partial_{x_\parallel} f(\vl x^\star) = \xi \sigma^{-\xi} p S_{d-2} \int_{-\Delta}^{\Delta} \mr{d}u (R + u)^{d-1} \int_0^\pi \mr{d}\phi \sin^{d-2}\phi \rho(R + u) \alpha(R + u) y(R + u) I(u, \phi),
\end{equation}
where the vector of integration norm is $r = R + u$ and its angle with respect to the north pole is $\phi$. All other angles simply integrate to the $(d-2)$-sphere surface, $S_{d-2}$, since they don't contribute to the integrand 
\begin{equation}
    I(u, \phi) = (x_1 - x_1^\star) \abs{\vl x - \vl x^\star}^{\xi -2} = a_0(\phi) + a_1(\phi) u + \mc O(u^2),
\end{equation}
with
\begin{equation}
    a_0(\phi) = \frac{1}{2R} \Big[2 R^2 (1 - \cos\phi)\Big]^{\xi/2} \hspace{10pt} \text{and} \hspace{10pt} a_1(\phi) = \Big[1 - \frac{\xi}{2}(1- \cos \phi)\Big] \Big[2 R^2 (1 - \cos\phi)\Big]^{\xi/2-1}.
\end{equation}
The leading order contribution $a_0$ vanishes because of the charge conservation (\eref{eq:sphere_charge_conservation}), so that the gradient reads
\begin{equation}
    \partial_{x_1} f(\vl x^\star) \sim p \int_{-\Delta}^{\Delta} \mr{d}u (R + u)^{d-1} \rho(R + u) \alpha(R + u) y(R + u) u \int_0^\pi \mr{d}\phi \sin^{d-2}\phi a_1(\phi) \sim p \Delta^2 \bar\alpha
\end{equation}
and the second scaling relation $p \bar\alpha \Delta^3 \sim 1$ is identical as for the stripe model. Since the other relations are obtained from local arguments, they are independent on the global shape of the classification task. The scaling laws for the spherical model are thus also given by \eref{eq:power_laws} and \eref{eq:scalelargesigma}.

\clearpage

\end{document}